\documentclass[final,5p,times,twocolumn]{elsarticle}
\usepackage{booktabs}       
\usepackage{mathtools}
\usepackage{amsmath}
\usepackage{subfig}
\usepackage[ruled,vlined,linesnumbered]{algorithm2e}
\newcommand{\parallelsum}{\mathbin{\|}}

\usepackage{amssymb}

\journal{Neural Networks}

\begin{document}

\begin{frontmatter}



\title{ Accelerating Deep Neural Network Training with Inconsistent Stochastic Gradient Descent }


\author[label1, label2]{Linnan Wang}
\ead{linnan.wang@gatech.edu}
\address[label1]{Georgia Institute of Technology}
\author[label2]{Yi Yang}
\author[label2]{Renqiang Min}
\author[label2]{Srimat Chakradhar}
\address[label2]{NEC Laboratories, USA }

\begin{abstract}
Stochastic Gradient Descent (SGD) updates network parameters with a noisy gradient computed from a random batch, and each batch evenly updates the network once in an epoch. This model applies the same training effort to each batch, but it overlooks the fact that the gradient variance, induced by Sampling Bias and Intrinsic Image Difference, renders different training dynamics on batches. In this paper, we develop a new training strategy for SGD, referred to as Inconsistent Stochastic Gradient Descent (ISGD) to address this problem. The core concept of ISGD is the inconsistent training, which dynamically adjusts the training effort w.r.t the loss. ISGD models the training as a stochastic process that gradually reduces down the mean of batch's loss, and it utilizes a dynamic upper control limit to identify a large loss batch on the fly. Then, it solves a new subproblem on the identified batch to accelerate the training while avoiding drastic parameter changes. ISGD is straightforward, computationally efficient and without requiring auxiliary memories. A series of empirical evaluations on real world datasets and networks demonstrate the promising performance of inconsistent training.
\end{abstract}


\end{frontmatter}


\section{Introduction}

The accessible TFLOPs brought forth by accelerator technologies bolster the booming development in Neural Networks. In particular, large scale neural networks have drastically improved various systems in natural language processing \cite{gimpel2010distributed}, video motion analysis \cite{donahue2015long}, and recommender systems \cite{yu2014parallel}. However, training a large neural network saturated with nonlinearity is notoriously difficult. For example, it takes 10000 CPU cores up to days to complete the training of a network with 1 billion parameters \cite{dean2012large}. Such computational challenges have manifested the importance of improving the efficiency of gradient based training algorithm.

The network training is an optimization problem that searches for optimal parameters to approximate the intended function defined over a finite training set. A notable aspect of training is the vast solution hyperspace defined by abundant network parameters. The recent ImageNet contests have seen the parameter size of Convolutional Neural Networks (CNN) increase to $n \sim 10^9$. Solving an optimization problem at this scale is prohibitive to the second order optimization methods, as the required Hessian matrix, of size $10^9\times10^9$, is too large to be tackled by modern computer architectures. Therefore, the first order gradient descent is widely used in training the large scale neural networks.

The standard first order full Gradient Descent (GD), which dates back to \cite{cauchy1847methode}, calculates the gradient with the whole dataset. Despite the appealing linear convergence rate of full gradient descent ($\mathcal{O}(\rho^{k}), \, \rho < 1$) \cite{roux2012stochastic}, the computation in an iteration linearly increases with the size of dataset. This makes the method unsuitable for neural networks trained with the sheer volume of labelled data. To address this issue, Stochastic Gradient Descent \cite{robbins1951stochastic, le2004large} was proposed by observing a large amount of redundancy among training examples. It approximates the dataset with a batch of random samples, and uses the stochastic gradient computed from the batch to update the model. Although the convergence rate of SGD, $\mathcal{O}({1/ \sqrt{bk}} + 1/k)$ \cite{bottou1998online} where $b$ is the batch size, is slower than GD, SGD updates the model much faster than GD in a period, i.e. larger $k$. As a result, the faster convergence is observable on SGD compared to GD in practice. SGD hits a sweet spot between the good system utilization \cite{wang2016blasx} and the fast gradient updates. Therefore, it soon becomes a popular and effective method to train large scale neural networks.

The key operation in SGD is to draw a random batch from the dataset. It is simple in math, while none-trivial to be implemented on a large-scale dataset such as ImageNet \cite{deng2009imagenet}. State of the art engineering approximation is the Fixed Cycle Pseudo Random (FCPR) sampling (defined in section \ref{FCPRS}), which retrieves batches from the pre-permuted dataset like a ring, e.g. $\mathbf{d_0} \rightarrow \mathbf{d_1} \rightarrow \mathbf{d_2} \rightarrow \mathbf{d_0} \rightarrow \mathbf{d_1} \rightarrow ... $ , where $\mathbf{d_i}$ denotes a batch. In this case, each batch receives the same training iterations as a batch updates the network exactly once in an epoch. Please note this engineering simplification allows batches to repetitively flow into the network, which is different from the random sampling in Statistics. However, it is known that the gradient variances differentiate batches in the training \cite{johnson2013accelerating}, and gradient updates from the large loss batch contribute more than the small loss ones \cite{simo2015discriminative}. This suggests that rebalancing the training effort across batches is necessary. SGD fails to consider the issue, and we think this is a problem.

In this paper, we propose Inconsistent Stochastic Gradient Descent (ISGD) to rebalance the training effort among batches. The inconsistency is reflected by the uneven gradient updates on batches. ISGD measures the training status of a batch by the associated loss. At any iteration $t$, ISGD traces the losses in iterations $[t-n_b, t]$, where $n_b$ is the number of distinct batches in a dataset. These losses assist in constructing a dynamic upper threshold to identify a under-trained batch during the training. If a batch's loss exceeds the threshold, ISGD accelerates the training on the batch by solving a new subproblem that minimizes the discrepancy between the loss of current batch and the mean. The subproblem also contains a conservative constraint to avoid overshooting by bounding the parameter change. The key idea of the subproblem is to allow additional gradient updates on a under-trained batch while still remaining the proximity to the current network parameters. Empirical experiments demonstrate ISGD, especially at the final stage, performs much better than the baseline method SGD on various mainstream datasets and networks.

For practical considerations, we also delve into the effect of batch size toward the convergence rate with system factors considered. Enlarging the batch size expedites the convergence \cite{byrd2012sample}, but it linearly adds computations in an iteration. In the scenario of single node training, 
a small batch is favored to ensure frequent gradient updates. In the scenario of the multi-node training, it entails heavy synchronizations among nodes per iteration. The more gradient updates, the higher synchronization cost is. In this case, a moderate large batch reduces overall communications \cite{li2014efficient}, and it also improves the system saturation and the available parallelism.

In summary, the novelties of this work are:
\vspace{-0.1in}
\begin{itemize}
  \item we propose a new training model, referred to as the inconsistent training, to improve the efficiency of SGD.
  \vspace{-0.1in}
  \item we apply the inconsistent training on SGD and its variants.
\end{itemize}

\section{Related Work}
A variety of approaches have been proposed to improve vanilla SGD for the neural network training. In this section, we demonstrate the concept of inconsistent training is fundamentally different from the existing methods.

The stochastic sampling in SGD introduces the gradient variance, which slows down the convergence rate \cite{bottou1998online}. The problem motivates researchers to apply various variance reduction techniques on SGD to improve the convergence rate. Stochastic Variance Reduced Gradient (SVRG) \cite{johnson2013accelerating} keeps network historical parameters and gradients to explicitly reduce the variance of update rule, but the authors indicate SVRG only works well for the fine-tuning of non-convex neural network. Chong et al. \cite{wang2013variance} explore the control variates on SGD, while Zhao and Tong \cite{zhao2014stochastic} explore the importance sampling. These variance reduction techniques, however, are rarely used in the large scale neural networks, as they consume the huge RAM space to store the intermediate variables. ISGD adjusts to the negative effect of gradient variances, and it does not construct auxiliary variables being much more memory efficient and practical than the variance reduction ones.

\textit{ Momentum } \cite{tseng1998incremental} is a widely recognized heuristic to boost SGD. SGD oscillates across the narrow ravine as the gradient always points to the other side instead of along the ravine toward the optimal. As a result, it tends to bounce around leading to the slow convergence. \textit{ Momentum } damps oscillations in directions of high curvature by combining gradients with opposite signs, and it builds up speed toward a direction that is consistent with the previously accumulated gradients \cite{sutskever2013importance}. The update rule of \textit{ Nesterov's accelerated gradient } is similar to \textit{ Momentum } \cite{sutskever2013training}, but the minor different update mechanism for building the velocity results in important behavior differences. Momentum strikes in the direction of the accumulated gradient plus the current gradient. In contrast, \textit{ Nesterov's accelerated gradient } strikes along the previous accumulated gradient, then it measures the gradient before making a correction. This prevents the update from descending fast, thereby increases the responsiveness. ISGD is fundamentally different from these approaches by considering the training dynamics on batches. ISGD rebalances the training effort across batches, while \textit{ Momentum } and \textit{ Nesterov's accelerated gradient } leverage the curvature tricks. Therefore, the inconsistent training is expected to be compatible with both methods.

\textit{ Adagrad } \cite{duchi2011adaptive} adapts the learning rate to the parameters, performing larger updates for infrequent parameters, and smaller updates for frequent parameters. It accumulates the squared gradients in the denominator, which will drastically shrink the learning rate. Subsequently, \textit{ RMSprop } and \textit{  Adadelta } have been developed to resolve the issue. These adaptive learning rate approaches adjust the extent of parameter updates w.r.t the parameter's update frequency to increase the robustness of training, while ISGD adjusts the frequency of a batch's gradient updates w.r.t the loss to improve the training efficiency. From this perspective, ISGD is different from the adaptive learning rate approaches.

The core concept of inconsistent training is to spare more training effort on the large loss batches than the small loss ones. The rational behind the scene is that gradient updates from the small loss batches contribute less than the large loss ones. Simo-Serra et al. \cite{simo2015discriminative} adopt a similar idea in training the Siamese network to learn the deep descriptors by intentionally feeding the network with hard training pairs, i.e. pairs yield large losses, and the method is proven to be an effective way to improve the performance. They manually pick the hard pairs to feed the network, while ISGD automatically identifies the hard batch during the training. In addition, the mechanism of ISGD's hard batch acceleration is different from the Simo-Serra's method. ISGD solves a sub-optimization problem on the hard batch to reduce the batch's loss and avoids drastic parameter changes, while the Simo-Serra's method simply feeds the batch more often. Please note it is important to bound the parameter changes, because overshooting a batch leads to the divergence on other batches. In summary, ISGD is the first neural network solver to consider the batch-wise training dynamics, and it has demonstrated promising performance on a variety of real world datasets and models.

\section{Problem Statement}
This section demonstrates the non-uniform batch-wise training dynamics. Theoretically, we prove the contribution of gradient updates varies among batches based on the analysis of SGD's convergence rate. We also hypothesize that Intrinsic Image Differences and Sampling Bias are high level factors to the phenomenon, and the hypothesis is verified by two controlled experiments. Both theories and experiments support our conclusion that the contribution of a batch's gradient update is different.

Then we demonstrate the Fixed Cycle Pseudo Random sampling employed by SGD is inefficient to handle this issue. In particular, the consistent gradient updates on all batches, regardless of their statuses, is wasteful especially at the end of training, and the gradient updates on the small loss batch could have been used to accelerate large loss batches.

\subsection{A Recap of CNN Training}
We formulate the CNN training as the following optimization problem. Let $\psi$ be a loss function with weight vector $\mathbf{w}$ as function parameters, which takes a batch of images $\mathbf{d}$ as the input. The objective of CNN training is to find a solution to the following optimization 
problem:
\begin{equation}
\label{CNN_Training}
\underset{\mathbf{w}}{min} \quad \psi_{\mathbf{w}}(\mathbf{d}) + \frac{1}{2}\lambda\parallelsum \mathbf{w} \parallelsum_2^2
\end{equation}
The second term is Weight Decay \cite{moody1995simple}, and $\lambda$ is a parameter to adjust its contribution (normally around $10^{-4}$). The purpose of Weight Decay is to penalize the large parameters so that static noise and irrelevant components of weight vectors get suppressed. \cite{moody1995simple}.

A typical training iteration of CNN consists of a Forward and Backward pass. Forward pass yields a loss that measures the discrepancy between the current predictions and the truth. Backward pass calculates the gradient, the negative of which points to the steepest descent direction. Gradient Descent updates the $\mathbf{w}$ as follows:
\begin{equation}
\mathbf{w}^{t} = \mathbf{w}^{t-1} - \eta_t \nabla \psi_{\mathbf{w}}(\mathbf{d}) \label{gd_update} 
\end{equation}
Whereas evaluating the gradient over the entire dataset is extremely expensive especially for large datasets such as ImageNet. To resolve this issue, mini-batched SGD is proposed to approximate the entire dataset with a small randomly drawn sample $\mathbf{d}_{t}$. The upside of mini-batched SGD is the efficiency of evaluating a small sample in the gradient calculation, while the downside is the stochastic gradient slowing down the convergence. Let's define a sample space $\Omega$. If $\psi_{\mathbf{w}}(\mathbf{d_{t}})$ is a random variable defined on a probability space ($\Omega, \Sigma, P$), the new objective function is
\begin{equation}
\label{sgd_objectives}
\underset{\mathbf{w}}{min} \quad E\{\psi_{\mathbf{w}}(\mathbf{d_{t}})\} = \int_\Omega \psi_{\mathbf{w}}(\mathbf{d_{t}})dP  + \frac{1}{2}\lambda\parallelsum \mathbf{w} \parallelsum_2^2
\end{equation}
the update rule changes to
\begin{equation} 
\mathbf{w}^{t} = \mathbf{w}^{t-1} - \eta_t \nabla \psi_{\mathbf{w}}(\mathbf{d_{t-1}}) \label{gd_update} 
\end{equation}
and the following holds, 
\begin{equation}
E\{\nabla \psi_{\mathbf{w}}(\mathbf{d_{t}}) \} = \nabla \psi_{\mathbf{w}}(\mathbf{d})
\end{equation}

\subsection{Measure Training Status with Cross Entropy Error}
\label{loss_status}
We use the loss to reflect the training status of a batch. A convolutional neural network is a function of $R^n \rightarrow R$, the last layer of which is a softmax loss function calculating the cross entropy between the true prediction probabilities $p(x)$ and the estimated prediction probabilities $\hat{p}(x)$. The definition of softmax loss function of a batch at iteration $t$ is 
\begin{equation} \label{weight_decay}
\psi_{\mathbf{w_{t}}}(\mathbf{d_{t}}) = -\sum\limits_{i}^{n_b}\sum\limits_{x}p(x)\log \hat{p}(x) + \frac{1}{2}\lambda\parallelsum \mathbf{w} \parallelsum_2^2
\end{equation}
where ${n_b}$ is the number of images in a batch, and $\lambda$ regulates Weight Decay. Since Weight Decay is applied, the loss of a batch fluctuates around a small number after being fully trained.

The loss produced by the cross entropy is a reliable indicator of a batch's training status. Given a batch $\mathbf{d_{t}}$, the cross entropy $\psi_{\mathbf{w_{t}}}(\mathbf{d_{t}})$ measures the discrepancy between the estimated probabilities and the truth. In the image classification task, the truth $p(x)$ is a normalized possibility vector, containing most zeros with only one scalar set to 1. The index of the vector corresponds to an object category. For example, $p(x) = [0, 0, 1, 0, 0]$ indicates the object belongs the category 2 (index starts from 0). The neural network produces an normalized estimate possibility $\hat{p}(x)$, and the loss function only captures the extent of making the correct prediction as the zeros in $p(x)$ offset the incorrect predictions in $\hat{p}(x)$. If $\hat{p}(x)$ is close to $p(x)$, the loss function yields a small value. If $\hat{p}(x)$ is far from $p(x)$, the loss function yields a large value. Therefore, we use the loss of a batch to assess the model's training status on it. Intuitively a large loss indicates that most predictions made by the network on the batch are false, and the additional training on the batch is necessary.

\begin{figure*}[t]
\centering
\subfloat[][single-class batches: $\mathbf{b_i}$ randomly draws 1000 images from a category of CIFAR-10.]{\includegraphics[width=0.9\textwidth]{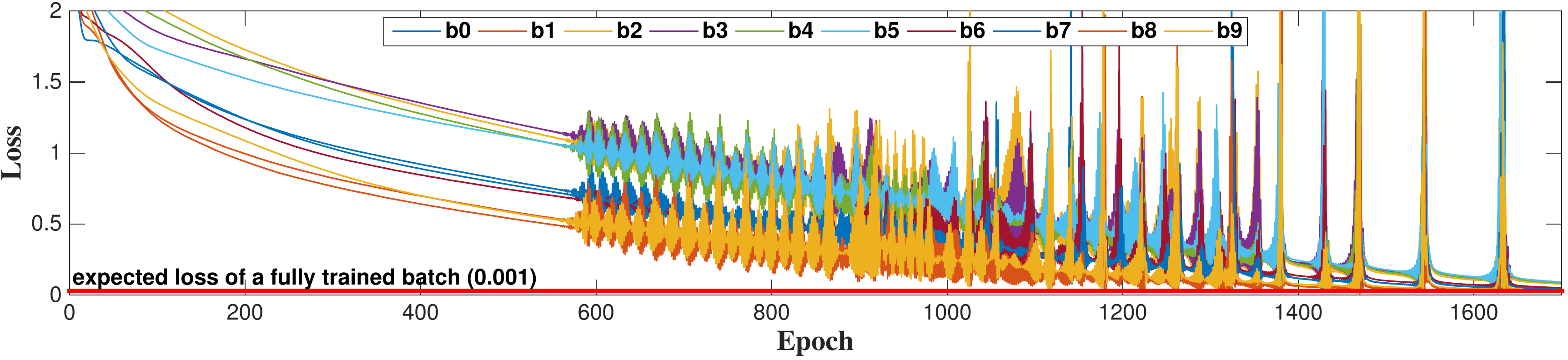}\label{sgd_loss_speed}} 
\\
\vspace{-0.15in}
\subfloat[][independent identically distributed (i.i.d) batches: $\mathbf{b_i}$ randomly draw 100 images from each categories of CIFAR-10 (10 categories in total). ]{\includegraphics[width=0.9\textwidth]{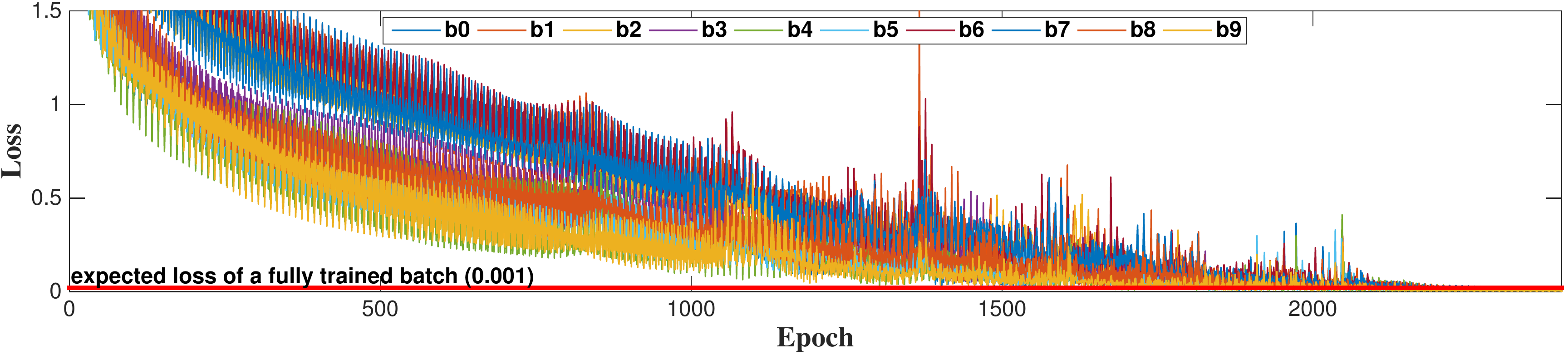}\label{uniform_batch_sgd_loss_speed}}
\caption{The loss traces of 10 single class and i.i.d  batches in two controlled experiments. We utilize an example network provided in Caffe. There is no cropping on input images and we shuffle the images within a batch. The datasets for each experiment is synthesized from CIFAR-10. }
\label{batch_loss_variation}
\end{figure*}

\subsection{ Motivation: Non-uniform Training Dynamics of Batches }

The gradient variance is the source of batch-wise training variations. The benefit of using a random sample to approximate the population is the significantly less computations in an iteration, while the downside is the noisy gradient. Please note the convergence rate in this section is measured by iterations. To analyze the training dynamics per iteration, we need define the Lyapunov process:
\begin{equation}
\label{lyapunov_process}
h_t = \parallel \mathbf{w}^{t} - \mathbf{w}^{*} \parallel_2^2 
\end{equation} 
The equation measures the distance between the current solution $\mathbf{w^t}$ and the optimal solution $\mathbf{w^{*}}$. $h_t$ is a random variable. Hence the convergence rate of SGD can be derived using 
Eq.\ref{gd_update} and Eq.\ref{lyapunov_process}:
\begin{equation}
\label{convergence_rate}
\begin{split}
h_{t+1} - h_t & = \parallel \mathbf{w}^{t+1} - \mathbf{w}^{*} \parallel_2^2 - \parallel \mathbf{w}^{t} - \mathbf{w}^{*} \parallel_2^2 \\
                     & = (\mathbf{w^{t+1}} + \mathbf{w^{t}} - 2\mathbf{w^{*}})(\mathbf{w^{t+1}} - \mathbf{w^{t}}) \\
                     & = (2\mathbf{w^{t}} - 2\mathbf{w^{*}} - \eta_t \nabla\psi_{\mathbf{w}}(\mathbf{d_{t}}))(-\eta_t \nabla\psi_{\mathbf{w}}(\mathbf{d_{t}})) \\
                     & = -2\eta_t(\mathbf{w}^t - \mathbf{w}^*)\nabla \psi_{\mathbf{w}}(\mathbf{d_{t}}) + \eta_t^2(\nabla\psi_{\mathbf{w}}(\mathbf{d_{t}}))^2 
\end{split}
\end{equation}
$\mathbf{d}_{t}$ is a random sample of $\mathbf{d}$ in the sample space $\Omega$, and $h_{t+1} - h_{t}$ is a random variable that depends on the drawn sample $\mathbf{d_{t}}$ and learning rate $\eta_t$. It suggests how far an iteration step toward $\mathbf{w}^{*}$. This equation demonstrates two important insights:

\begin{itemize}
    \item {\textit{Reducing $\mathbf{VAR}\{ {\nabla\psi_{\mathbf{w}}(\mathbf{d_{t}})} \} $ improves the convergence rate}}. The expectation of Eq.\ref{convergence_rate} yields the average convergence rate at the precision of an iteration.
\begin{equation} \label{convergence_rate_exp}
\begin{split}
\mathbf{E}\{ h_{t+1} - h_t \} &= -2\eta_t(\mathbf{w}^t - \mathbf{w}^*)   \mathbf{E}\{\nabla\psi_{\mathbf{w}}(\mathbf{d_{t}}) \}    \\
                                             &+ \eta_t^2\mathbf{E}\{ (\nabla\psi_{\mathbf{w}}(\mathbf{d_{t}}))^2 \}                                          \\
                                            &= -2\eta_t(\mathbf{w}^t - \mathbf{w}^*)   \mathbf{E}\{\nabla\psi_{\mathbf{w}}(\mathbf{d_{t}})\}    \\
              &+ \eta_t^2(\mathbf{E}\{ \nabla\psi_{\mathbf{w}}(\mathbf{d_{t}}) \})^2 + \mathbf{VAR}\{ {\nabla\psi_{\mathbf{w}}(\mathbf{d_{t}})} \} \\
\end{split}
\end{equation}
To simplify the analysis of Eq.\ref{convergence_rate_exp}, let's assume the convexity on $\psi_{\mathbf{w}}(\mathbf{d_{t}})$ implying that 
\begin{equation}
h_{t+1} - h_t < 0
\end{equation}
\begin{equation}
-(\mathbf{w}^t - \mathbf{w}^*)\mathbf{E}\{\nabla\psi_{\mathbf{w}}(\mathbf{d_{t}})\} < 0. 
\end{equation}
     $\mathbf{E} \{ \nabla\psi_{\mathbf{w}}(\mathbf{d_{t}}) \}$ is the unbiased estimate of $\mathbf{E} \{ \nabla\psi_{\mathbf{w}}(\mathbf{d}) \}$. Therefore, maximizing the contribution of an iteration is reduced to the minimization of $\mathbf{VAR}\{ {\nabla\psi_{\mathbf{w}}(\mathbf{d_{t}})} \}$. This direction has been well addressed \cite{reddi2015variance, reddi2016stochastic}.

   \item {\textit{The contribution of an iteration, $h_{t+1} - h_t$, varies with respect to $\mathbf{d_{t}}$.}} According to Eq.\ref{convergence_rate}, the variance of $h_{t+1} - h_t$ is:
   \begin{equation} \label{convergence_rate_var}
\begin{split}
\mathbf{VAR}\{ h_{t+1} - h_t \} = 4\eta_t^2(\mathbf{w}^t - \mathbf{w}^*)^2 \mathbf{VAR}\{\nabla \psi_{\mathbf{w}}(\mathbf{d_{t}}) \} \\ 
+ \eta_t^4 \mathbf{VAR}\{ (\nabla\psi_{\mathbf{w}}(\mathbf{d_{t}}))^2 \} \\
 - 2\eta_t^3(\mathbf{w}_t - \mathbf{w}_{*})\mathbf{COV}\{\nabla \psi_{\mathbf{w}}(\mathbf{d_{t}}), \nabla \psi_{\mathbf{w}}(\mathbf{d_{t}})^2 \} \\
\end{split}
\end{equation}
The equation demonstrates $\mathbf{VAR}\{ h_{t+1} - h_t \}  \neq 0$, which implies the contribution of gradient updates is non-uniform. It is interesting to notice that the determining factors in this equation, $\nabla \psi_{\mathbf{w}}(\mathbf{d_{t}})^2$ and $\nabla \psi_{\mathbf{w}}(\mathbf{d_{t}})$, is contingent upon $\mathbf{d_{t}}$, suggesting a correlation between $h_{t+1} - h_{t}$ and $\mathbf{d_{t}}$. This unique insight motivates us to understand what factors in $\mathbf{d_{t}}$ affect the convergence rate $h_{t+1} - h_{t}$, and how to address the load balancing problem in the training. Although there are extensive studies toward the variance reduction on $\nabla \psi_{\mathbf{w}}(\mathbf{d_{t}})$, few explores this direction. Let's use the loss of a batch to measure the model's training status on it (explained in section \ref{loss_status}). 
Fig.\ref{batch_loss_variation} demonstrates the loss traces of 10 separate batches during the training. It is observable that  the losses of batches degenerate at different rates. Therefore, the empirical observations and Eq.\ref{convergence_rate_var} prompt us to conclude that
\begin{center}
\vspace{-0.1in}
 \textit{the contribution of a batch's gradient update is non-uniform}
\vspace{-0.1in}
\end{center}
This also explains the distinctive training dynamics of batches in Fig.\ref{batch_loss_variation}. Eq.\ref{convergence_rate_var} suggests $\mathbf{d}_t$ is critical for the claim. We conduct a set of empirical evaluations to understand how $\mathbf{d}_t$ affect $\mathbf{VAR}\{ h_{t+1} - h_t \}$, and we propose two high level factors, Sampling Bias and Intrinsic Image Difference, to explain the phenomenon. The definitions of these two terms are as follows:\\
\textbf{\textit{Sampling Bias}}: It is a bias in which a sample is collected in such a way that some members of the intended population are less likely to be included than others. \\
\textbf{\textit{Intrinsic Image Difference}}: Intrinsic Image Difference indicates images from the same subpopulation are also different at pixels. For example, the category 'cat' can contain some white cat pictures or black cat pictures. Though black cat and white cat belong to the cat subpopulation, they are still different at pixels. 
\end{itemize}

Sampling Bias is the first factor to explain the training variations on batches. We consider two kinds of Sampling Bias. First, existing datasets, such as Places \cite{zhou2014learning} or ImageNet, contain uneven number of images in each category. As a result, the dominate sub-population is more likely to be selected in a batch than others. Second, the insufficient shuffling on the dataset may lead to clusters of subpopulations. When SGD sequentially draws images from the insufficient permuted dataset to form a randomized batch (explained in section.\ref{FCPRS}), one subpopulation is more likely to be included than others. In both cases, they conform to the definition of Sampling Bias. For example, the chance of sampling $1$ from [1, 1, 1, 0, 2, 3] is higher than the rest. To support the claim, we synthesized 10 single-class batches randomly drawn from an exclusive image category in CIFAR-10 \cite{krizhevsky2010convolutional}. Please note CIFAR-10 contains 10 independent image categories. Each batch represents a unique CIFAR-10 category, and they are highly polluted with Sampling Bias as each batch only contains one subpopulation. Fig.\ref{sgd_loss_speed} demonstrates the loss traces of ten single-class batches. It is obvious to see the losses of ten batches degrade independently. In particular, gradient updates from the yellow batch is more effective than the purple batch. Therefore, these results justify our claim about Sampling Bias and the batch-wise training variation.

Intrinsic Image Difference is the second factor to explain the training variations on batches. To substantiate this point, we conduct a controlled experiment on 10 independent identically distributed batches. A batch includes 1000 images, and each batch contains 100 randomly drawn images from category 0, 100 images from category 1, ... , 100 images from category 9. This sequence is fixed across batches to eliminate the potential ordering influence. In this case, each batch contains the same number of images from 10 subpopulations in CIFAR-10 and the only difference among them is the pixels. Hence, we consider these batches independent identically distributed. The network is same as the one used in Sampling Bias. Fig.\ref{uniform_batch_sgd_loss_speed} demonstrates the loss traces of 10 i.i.d batches. Although a strong correlation persists through the training, it is still clear the losses of i.i.d batches degrade at separate rates. Particularly, the loss of batch 4 (green) is around 0.5 while batch 3 (purple) is around 1.3 at the epoch 400. Please note these batches are i.i.d, and they are supposed to be approximately identical to the original dataset. However, the training variations still exist indicating the non-uniform contribution of gradient updates from each batches.

\subsection{Problems of Consistent Training in SGD}
\label{FCPRS}
SGD relies on a key operation, uniformly drawing a batch from the entire dataset. It is simple in math but 
nontrivial in the system implementation. ImageNet, ILSVRC2012 for example, contains 1431167 256$\times$256 high resolution RGB images accounting for approximately 256 GB in total size. Uniformly drawing a random batch from the 256 GB binary file involves significant overhead 
such as TLB misses or random Disk I/O operations. In addition, the drastic speed gap between Processor and Disk further deteriorates the issue. Existing deep learning frameworks, such as Caffe \cite{jia2014caffe} or Torch \cite{collobert2011torch7}, alleviates the issue by pre-permuting the 
entire dataset before slicing into batches:
$ Permute\{ \mathbf{d} \} \rightarrow \mathbf{d} = \{\mathbf{d_0},\mathbf{d_1}, ... ,\mathbf{d_{n-1}},\mathbf{d_n} \}  = \Omega $. 
During the training, each iteration fetches a batch from the permuted dataset $\Omega$ in a sequential manner $\mathbf{d_0} \rightarrow \mathbf{d_1} \rightarrow ... \rightarrow \mathbf{d_n}$; and restart fetching from the beginning $\mathbf{d_0}$ after $\mathbf{d_n}$, creating a fixed circle batch retrieval pattern. We refer to this sampling method as Fixed Circle Pseudo Random Sampling. The random reads are subsequently reduced to sequential reads on Disk. Therefore, FCPR Sampling is widely adopted by SGD. Let $n_{\mathbf{d}}$ to be the size of a dataset and $n_{\mathbf{b}}$ to be the batch size. The size of sample space is $n_{\mathbf{d}}$/ $n_{\mathbf{b}}$, and the batch being assigned to iteration $j$ is $\mathbf{d_{t}}$, where
\begin{displaymath}
 t = j \mbox{ } \mbox{mod} * {\frac{n_{\mathbf{d}}}{ n_{\mathbf{b}}}}
\end{displaymath}
At any iteration, the model always anticipate a fixed batch, as the batch will flow into the model at iteration $t+1 \cdot epoch$, ..., $t + n \cdot epcoh$.
If the training of a batch is dominated by the gradient update on itself, the loss of this batch is predominately reduced at iteration $t, t+1*epoch, t+2*epoch, ..., t+n*epoch$. Since the contribution from a batch's gradient update is different, the repetitive batch retrieval pattern fosters the batches' distinctive training speeds. However, the FCPR sampling in SGD treats batches identically. 

The problem of FCPR sampling is the consistent gradient updates on batches regardless of the model's training status. It is inefficient to update a small loss batch as frequently as a large loss batch. Fig.\ref{uniform_batch_sgd_loss_speed} demonstrates the yellow batch are fully trained after the epoch 1600, while the blue batch does not until the epoch 2100. During epochs [1600, 2100], the yellow batch stays fully trained most of time indicating unnecessary training iterations on it. Besides, we also verify that the contribution of a batch's gradient update is different. Therefore, regulating the training iterations w.r.t the model's training status on batches will improve the efficiency of SGD.

\section{Inconsistent Stochastic Gradient Descent}

In this section, we present Inconsistent Stochastic Gradient Descent that rebalances the training effort w.r.t a batch's training status. The inconsistency is reflected by the non-uniform gradient updates on batches. The first question is how to dynamically identify a slow or under-trained batch during the training. We model the training as a stochastic process, and apply the upper control limit to dynamically identify a under-trained batch. The second question is how to accelerate a under-trained batch. We propose a new optimization to be solved on the batch, the objective of which is to accelerate the training without drastic parameters changes. For practical considerations, we also study the effects of ISGD batch size on the convergence rate, system saturations and synchronization cost.

\subsection{Identifying Under-trained Batch}

\begin{figure}[!t]
\centering
\includegraphics[height=1.5in]{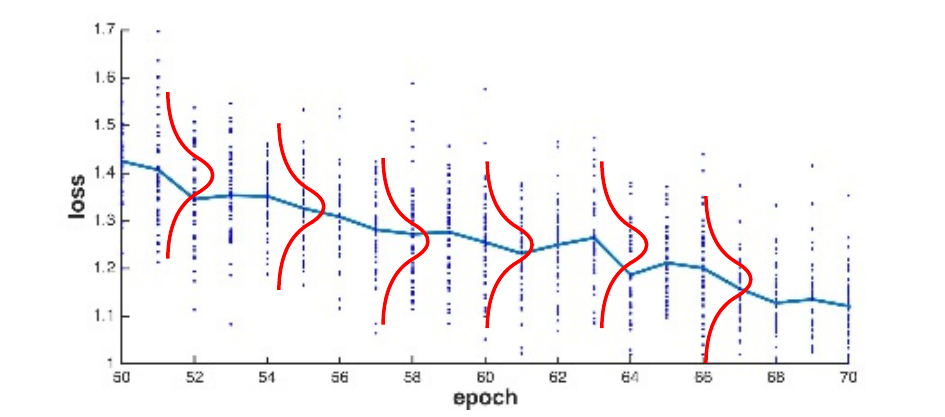}
\caption{Each blue dots depicts the loss of a batch produced from the CIFAR network in Caffe on CIFAR-10, and losses are grouped by epochs. The central line is the average loss. From the figure, it is legitimate to assume the losses of batches in an epoch follow the normal distribution, and the training is at reducing the mean of population tile the network converges.}
\label{norm_dist}
\end{figure}

ISGD models the training as a stochastic process that slowly reduces down the average loss of batches. We assume the normal distribution on the batch's loss in an epoch. The reasons are that: 1) SGD demands a small learning rate ($lr$) \cite{bottou2010large} to converge, and $lr$ is usually less than $10^{-1}$. $lr$ determines the step length, while the normalized gradient determines the step direction. The small value of $lr$ limits the contribution made by a gradient update, thus the training process is to gradually reduce down the loss toward a local optimal.  2) Each batch represents the original dataset, and there exists a strong correlation among batches in the training. This implies that the loss of a batch will not be drastically different from the average at any iteration t.
Fig.\ref{norm_dist} demonstrates the loss distribution of training a network on CIFAR-10, in which the losses are arranged by epochs. From the figure, it is valid to assume the normal distribution on the loss in an epoch. Therefore, we conclude that
\begin{center}
\textit{ the training is a stochastic process that slowly decreases the mean of losses tile the network converges. }  \\
\end{center}

The $3\sigma$ control limit is an effective method to monitor the abnormalities in a statistical process. Since we treat the training as a process that decreases the average loss of batches, ISGD utilizes the upper control limit to dynamically identify abnormal large-loss batches on the fly. To get the limit, ISGD calculates two important descriptive statistics, the running average loss $\bar{\psi}$ and the running standard deviation $\sigma_\psi$ during the training. ISGD keeps a queue to store the losses produced by iterations in $[t-n_b, t]$, where $n_b$ is the size of sample space (or the number of batches in an epoch). The queue functions as a moving window tracking the loss information in the previous epoch to yield $\bar{\psi}$ and $\sigma_\psi$. 
\begin{equation}
\label{average_loss}
\bar{\psi}  =  \frac{1}{n_b} \sum\limits_{i=1}^{n_b} \psi_\mathbf{w_{t-i}}(\mathbf{d_{t-i}})
\end{equation}
\begin{equation}
\sigma_\psi =  \sqrt{ \frac{1}{n_b} \sum\limits_{i=1}^{n_b} [\psi_\mathbf{w_{t-i}}(\mathbf{d_{t-i}}) - \bar{\psi}]^2 }
\end{equation}
Since the queue length is fixed at $n_b$ and the loss of a batch is a float number, the calculations of $\bar{\psi}$ and $\sigma_\psi$ and the memory cost for the queue are O(1) at any iteration $t$. Therefore, ISGD is much more memory efficient than the variance reduction approaches that require intermediate variables of the same size as the network parameters. With $\bar{\psi}$ and $\sigma_\psi$, the upper control limit is
\begin{equation}
\label{control_limits}
limit = \bar{\psi} + 3\sigma_\psi
\end{equation}
In this case, we adopt the $3\sigma$ control limit. The multiplier before the $\sigma_\psi$ plays an important role in between the exploration of new batches and the exploitation of the current batch. Please refer to the discussion of Alg.\ref{ugd_alg} in section \ref{additional_discussion_isgd} for more discussion. If the loss of current iteration $t$ is
\begin{equation}
\label{upper_limit}
\psi_\mathbf{w_{t-1}}(\mathbf{d_t}) > limit
\end{equation}
we consider $\mathbf{d_t}$ as a under-trained batch. 

\begin{figure}[!t]
\centering
\includegraphics[height=1.5in]{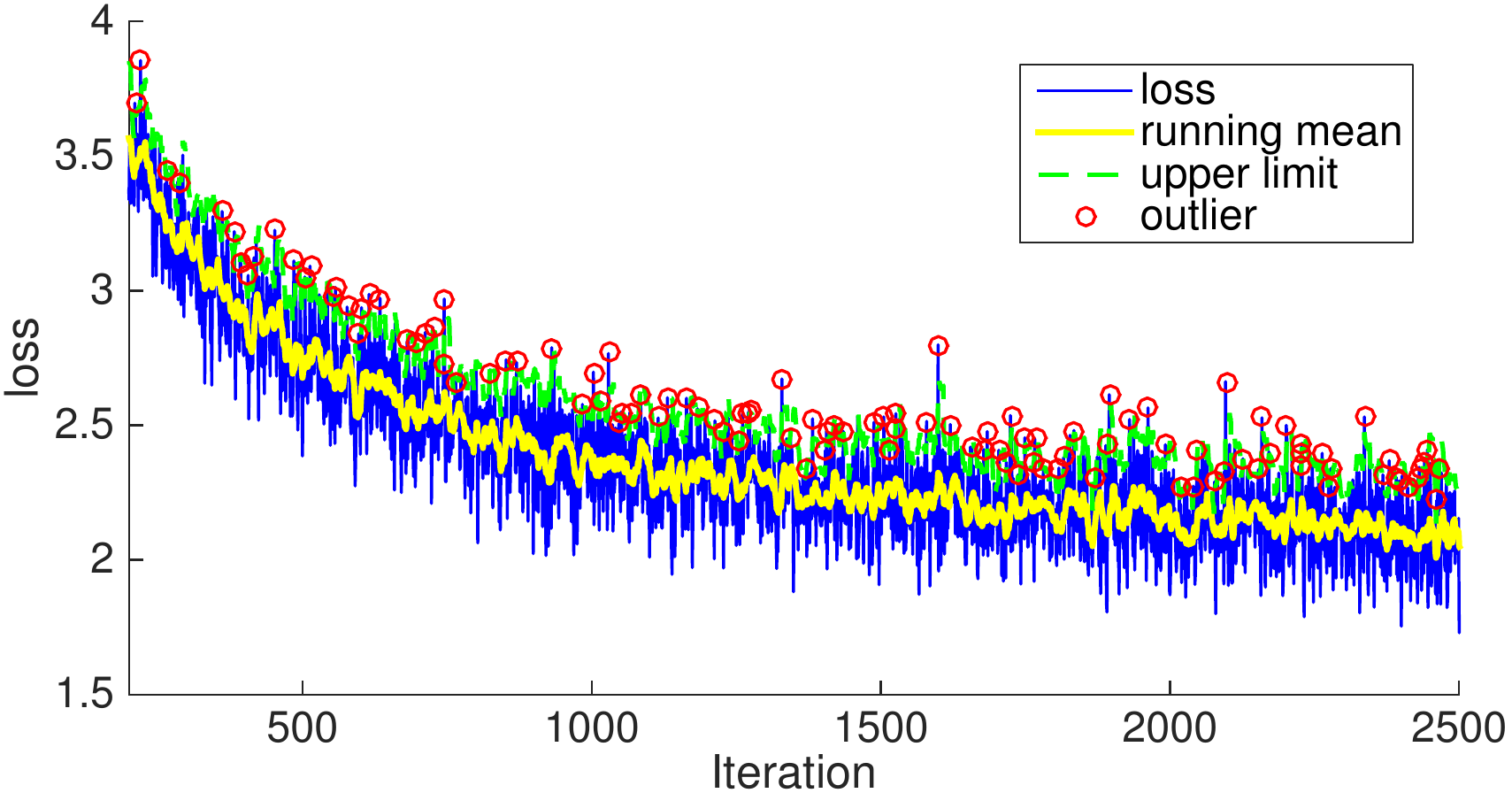}
\caption{Supervise the training of AlexNet with the proposed control limit. The red outliers are under-trained batches. ISGD subsequently accelerate the batches by solving Alg.\ref{chase_alg}.}
\label{control_chart}
\end{figure}

Fig.\ref{control_chart} demonstrates an example of the proposed method to identify a under-trained batch on the fly. The blue line is the loss of batch,  and the yellow line is the running average $\bar{\psi}$. The green line is the upper control limit, and the red dots are outliers considered as under-trained. The experiment is conducted with AlexNet on ImageNet, and it is clear that ISGD successfully identifies the large-loss batches in the training with the proposed approach.

\begin{algorithm}[!t]
\caption{Inconsistent Stochastic Gradient Descent}
\label{ugd_alg}
\DontPrintSemicolon
\KwData{$\mathbf{d} \mbox{ , } \mathbf{w}$ }
\KwResult{$\mathbf{w^\prime}$}
\Begin{
        iter = 0                                  \\
        $n = \frac{n_{training}}{n_{batch}}$         \\
        $\overline{\psi} = 0$                     \\
        $\sigma_\psi = 0$                         \\
        $limit = 0$                               \\
        $loss_{queue} \leftarrow \emptyset$       \\
        \While{$ not \mbox{ } converge $} {
            $broadcast( \mathbf{w} )$    \\
            $[\psi, \nabla\psi] = ForwardBackward(\mathbf{d_t})$                   \\
            $reduce(\psi)$  \\
            $reduce(\nabla\psi)$  \\
            \If{$\mbox{iter} < n$} {
                $loss_{queue}.push$($\psi$)               \\
                $\overline{\psi}  = \frac{ \overline{\psi} \cdot iter + \psi }{iter+1} $                    \\
            } \Else {
                $l = loss_{queue}.dequeue$()        \\
                $\sigma_{\psi} = STD(loss_{queue})$ \\
                $\overline{\psi}  = \frac{ \overline{\psi} \cdot n - l + \psi }{n} $   \\
                $limit = \overline{\psi} + 3*\sigma_{\psi}$
            }
            $\mathbf{w} = \mathbf{w} - \eta \cdot  \nabla\psi$ \\
            \If{$\psi > limit \enspace and \enspace iter > n$} {                 
                minimize Eq.\ref{new_optimization}  with Alg.2 \\
            }
            iter++
        }
}
\end{algorithm}

\subsection{Inconsistent Training}
The core concept of our training model is to spend more iterations on the large-loss batches than the small-loss ones. The batch retrieval pattern in ISGD is similar to FCPR sampling but with the following important difference. Once a batch is identified as under-trained, ISGD stays on the batch to solve a new sub-optimization problem to accelerate the training, and the batch receives additional training iterations inside the sub-problem. In this case, ISGD does not compromise the system efficiency of FCPR sampling, while it still regulates the training effort across the batches. The new subproblem is 

\begin{equation}
\begin{split}
\label{new_optimization}
\underset{\mathbf{w}}{min} \quad \phi_{\mathbf{w}}(\mathbf{d_t}) = & \frac{1}{2}\parallelsum \psi_{\mathbf{w}}(\mathbf{d_t}) - limit \parallelsum_2^2  \\
                                                                   & + \frac{\epsilon}{2n_w}\parallelsum \mathbf{w} - \mathbf{w_{t-1}}  \parallelsum_2^2
\end{split}
\end{equation}
where $n_w$ is the number of weight parameters in the network and $\epsilon$ is a parameter for the second term. The first term minimizes the difference between the loss of current under-trained batch $\mathbf{d_t}$ and the control limit. This is to achieve the acceleration effect. The second term is a conservative constraint to avoid drastic parameter changes. Please note the second term is critical because overshooting a batch negatively affects the training on other batches. The parameter $\epsilon$ adjusts the conservative constrain, and it is recommended to be $~10^{-1}$.
The derivative of Eq.\ref{new_optimization} is:
\begin{equation} \label{new_optimization_derivative}
\begin{split}
\nabla \phi_{\mathbf{w}}(\mathbf{d_t}) = & [\psi_{\mathbf{w}}(\mathbf{d_t}) - limit]\nabla\psi_{\mathbf{w}}(\mathbf{d_t}) \\
                                         &+ \frac{\epsilon(\mathbf{w} - \mathbf{w_{t-1}})}{n_w}                                     \\
\end{split}
\end{equation}
Please note $limit$, $\mathbf{w_{t-1}}$ and $\mathbf{d_t}$ are constants. Solving Eq.\ref{new_optimization} precisely incurs the significant computation and communication overhead, which offsets the benefit of it. In practice, we approximate the solution to the new subproblem, Eq.\ref{new_optimization}, with the early stopping. This avoids the huge searching time wasted on hovering around the optimal solution. A few iterations, 5 for example, are good enough to achieve the acceleration effects. Therefore, we recommend approximating the solution by the early stopping.  

\begin{algorithm}[!t]
\caption{Solving the conservative subproblem to accelerate a under-trained batch}
\label{chase_alg}
\DontPrintSemicolon
\KwData{$\mathbf{d_t} \mbox{, } \mathbf{w} \mbox{, }  \mathbf{w_{t-1}} \mbox{, } stop \mbox{, } limit$ }
\KwResult{$\mathbf{w}$}
\Begin{
        $iter$ = 0                                                                      \\
        \While{$ iter <  stop$ and $\psi > limit$} {
            $[\psi, \nabla\psi] = ForwardBackward(\mathbf{d_t})$                   \\
            $reduce(\psi)$                      \\
            $reduce(\nabla\psi)$                \\
            $\mathbf{w} = \mathbf{w} - \zeta\{[\psi_{\mathbf{w}}(\mathbf{d_t}) - limit]\nabla\psi_{\mathbf{w}}(\mathbf{d_t}) + \frac{\epsilon(\mathbf{w} - \mathbf{w_{t-1}})}{n_w}\} $ \\
            $broadcast( \mathbf{w})$    \\
            iter++
        }
}
\end{algorithm}

Alg.\ref{ugd_alg} demonstrates the basic procedures of ISGD. Since the training status of a batch is measured by the loss, ISGD identifies a batch as under-trained if the loss is larger than the control limit $\overline{\psi} + 3*\sigma_{\psi}$ (Line 20). A stringent limit triggers Eq.\ref{new_optimization} more frequently. This increases exploitation of a batch, but it also decreases the exploration of batches to a network in the fixed time. Therefore, a tight limit is also not desired.  A soft margin, 2 or 3 $\sigma_{\psi}$, is preferred in practice; and this margin is also widely applied in Statistical Process Control to detect abnormalities in a process. We recommend users adjusting the margin according to the specific problem. ISGD adopts a loss queue to dynamically track the losses in an epoch so that the average loss, $\overline{\psi}$, is calculated in O(1) (line 17). The loss queue tracks iterations in the previous epoch; the length of it equals to the length of an epoch. Similiarly, calculating $\sigma_{\psi}$ is also in O(1) time (line 18). We do not initiate Alg.\ref{chase_alg} until the first epoch to build up a reliable limit (line 22 the condition of iter $>$ n).

\label{additional_discussion_isgd}
Alg.\ref{chase_alg} outlines the procedures to solve the conservative subproblem on a under-trained batch. The conservative subproblem intends to accelerate the under-trained batch without drastic weight changes. The update equation in line 7 corresponds to Eq.\ref{new_optimization_derivative}. 
Specifically, $[\psi_{\mathbf{w}}(\mathbf{d_t}) - limit]\nabla\psi_{\mathbf{w}}(\mathbf{d_t})$ is the gradient of  $\frac{1}{2}\parallelsum \psi_{\mathbf{w}}(\mathbf{d_t}) - limit \parallelsum_2^2$ to accelerate the training of a under-trained batch; the second term, $\frac{\epsilon}{n_w}(\mathbf{w} - \mathbf{w_{t-1}})$, is the gradient of $\frac{\epsilon}{2n_w} \parallelsum \mathbf{w} - \mathbf{w_{t-1}}  \parallelsum_2^2$ that bounds significant weight changes. The $limit$ is the same upper control threshold in Alg.\ref{ugd_alg}. The $stop$ specifies the maximal approximate iterations to reflect the early stopping. $\zeta$ is a constant learning rate.

The neural network training needs gradually decrease the learning rate to ensure the convergence \cite{bottou1998online}. It is a common tactic to decrease the learning rate w.r.t training iterations. The inconsistent iterations of ISGD requires a new way to guide the learning rate. Instead, ISGD decreases the learning rate w.r.t the average loss of a dataset. The average loss is better than iterations, as it directly reflects the training status of the model, while calculating the average loss of a dataset is expensive. Since the average loss in Eq.\ref{average_loss} is from the latest scan of dataset (or losses in an epoch), it is approximate to the average loss of dataset. Hence, we use the average loss (Alg.\ref{ugd_alg}, line 19) to guide the learning rate.

\subsection{Extend to Other SGD Variants}
\label{sgd_variants}
It is straight forward to extend the inconsistent training to other SGD variants. For example, Momentum \cite{qian1999momentum} updates the weight with the following equations
\begin{equation}
\begin{split}
\label{momentum_update}
\mathbf{v_{t+1}}  &=  \mu \mathbf{v_{t}} - \alpha \nabla \psi({\mathbf{w_{t}}}) \\
\mathbf{w_{t+1}} &= \mathbf{w_{t}} + \mathbf{v_{t+1}}
\end{split}
\end{equation}
and the Nesterov accelerated gradient follows the update rule of
\begin{equation}
\begin{split}
\label{nesterov_update}
\mathbf{v_{t+1}}  &=  \mu \mathbf{v_{t}} - \alpha \nabla \psi({\mathbf{w_{t}} + \mu \mathbf{v_{t}}}) \\
\mathbf{w_{t+1}} &= \mathbf{w_{t}} + \mathbf{v_{t+1}}
\end{split}
\end{equation}
To introduce the inconsistent training to these SGD variants, we only need change the line 21 of Alg.\ref{ugd_alg} according to Eq.\ref{momentum_update} and Eq.\ref{nesterov_update}, respectively. The Alg.\ref{chase_alg} remains the same.

\subsection{Parallel ISGD}
\begin{figure}[!t]
\centering
\includegraphics[height=1.5in]{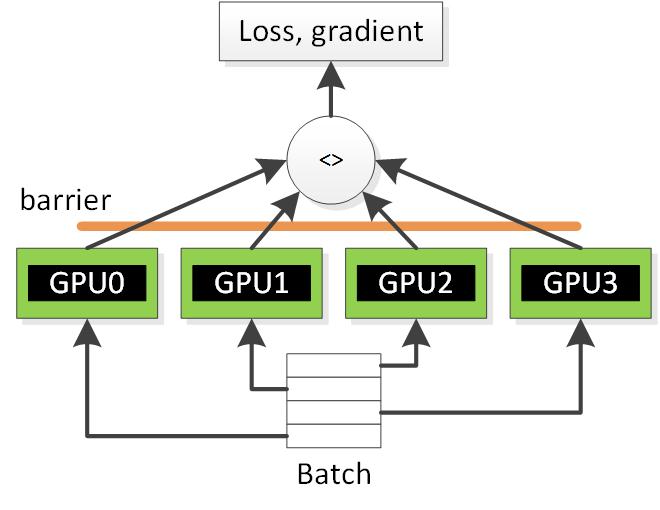}
\caption{the parallelization of ISGD with the data parallelism, and $<>$ indicates the $reduce$ of all sub-batches' gradients to yield the gradient of original batch.}
\label{parallel_isgd}
\end{figure}

ISGD intends to scale over the distributed or multiGPU system using MPI-style collectives such as $broadcast$, $reduce$, and $allreduce$ \cite{wang2016efficient}. Alg.\ref{ugd_alg} and Alg.\ref{chase_alg} are already the parallel version manifested by the collectives in them. Fig.\ref{parallel_isgd} demonstrates the data parallelization scheme inside ISGD. Let's assume there are $n$ computing nodes, each of which is a GPU or a server in a cluster. Each node contains a model duplicate. A node fetches an independent segment of the original batch
referred to as the sub-batch. Subsequently, all nodes simultaneously calculate sub-gradients and sub-losses with the assigned sub-batches. Once the calculation is done, the algorithm $reduce$ sub-gradients and sub-losses (Line 10-12 in Alg.\ref{ugd_alg}) to a master node so as to acquire a global gradient and loss. Then, the master node updates network weights (line 21 in Alg.\ref{ugd_alg}) and $broadcast$ (line 9 in Alg.\ref{ugd_alg}) the latest weights. Please refer to \cite{wang2016efficient} for more discussion regarding the MPI-style collectives and the SGD parallelization. Therefore, ISGD separates the algorithm from the system configurations by employing MPI-style collectives \cite{gabriel2004open}. Since MPI is an industrial and academia standard, ISGD is highly portable on various heterogeneous distributed system. 

\subsection{Batch Size and Convergence Speed}
Batch size is the key factor to the parallelism of ISGD. As operations on a batch are independent, scaling ISGD on systems with the massive computing power prefers a sufficiently large batch. An unwieldy large batch size, however, is detrimental to the convergence rate under the limited computing budget. Current convergence rate analysis utilizes iterations as the only performance metric, but it fails to consider the fact that an iteration faster algorithm may cost more time than the slower counterpart. Hence, it is practical to analyze the convergence rate in the time domain.

\begin{figure}[!t]
\centering
\includegraphics[width=3in]{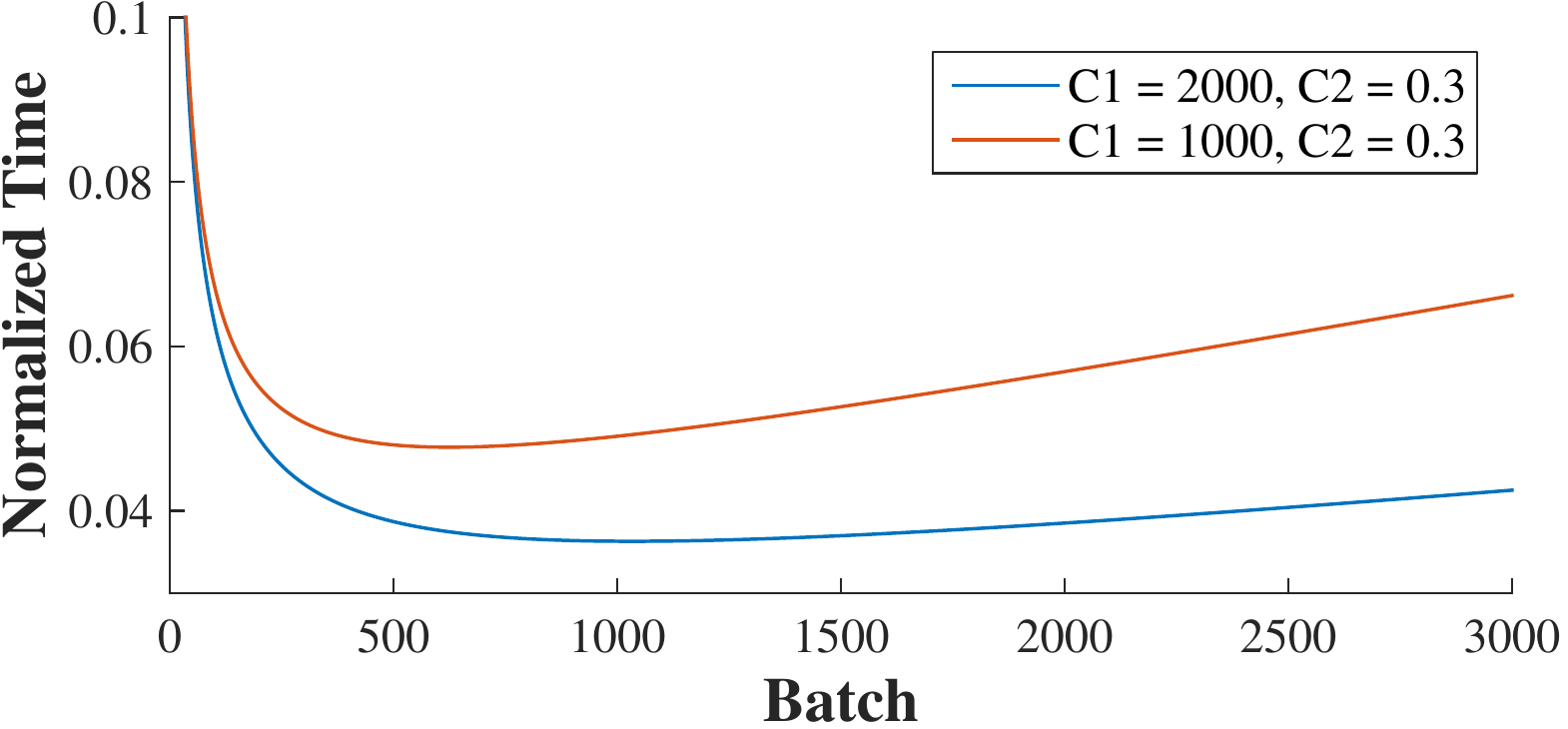}
\caption{The predicted training time calculated by Eq.\ref{loss_by_time} at different batch sizes.}
\label{convergence_rate_by_time}
\end{figure}

Let's assume the maximal processing capability of a system is $C_1$ images per second, and the time spent on synchronizations is $C_2$ seconds. Network cost is a constant because it only depends on the size of network parameter. A gradient update essentially costs:
\begin{equation}
\begin{split}
\label{performance}
t_{iter} & = t_{comp} + t_{comm} \\
           & = \frac{n_b}{C_1} + C_2
\end{split}
\end{equation}
where $n_b$ is the batch size. Given fixed time $t$, the number of gradient updates is
\begin{equation}
\label{iterations_time}
T = \frac{t}{t_{iter}}
\end{equation}
After T gradient updates, the loss is bounded by \cite{dekel2012optimal}
\begin{equation}
\label{loss_bound}
\psi \leqslant \frac{1}{\sqrt{n_bT}} + \frac{1}{T}
\end{equation}
Let's assume equality in Eq.\ref{loss_bound} and substitute Eq.\ref{iterations_time}. It yields Eq.\ref{loss_by_time} that governs loss $\psi$, 
time $t$ and system configurations $C_1$ and $C_2$:
\begin{equation}
\label{loss_by_time}
\psi t = \sqrt{t}\sqrt{\frac{n_b+C_1C_2}{n_bC_1}} + \frac{n_b}{C_1} + C_2
\end{equation}
Fig.\ref{convergence_rate_by_time} presents the predicted training time under two system configurations calculated by Eq.\ref{loss_by_time} at different batch sizes $n_b \in (0, 3000)$. By fixing $\psi$, the equation approximates the total training time under different batches. The figure demonstrates the optimal batch size of the first and second system are 500 and 1000 respectively. In this case, a faster system needs a larger batch. The performance of both systems deteriorates afterward. As a result, the optimal batch size is a tradeoff between system configurations and algorithmic convergences.

\begin{figure*}[t]
\vspace{-0.3in}
\centering
\subfloat[][ISGD loss distribution by epochs. The X and Y axis is same as (b) ]{\includegraphics[width=0.48\textwidth]{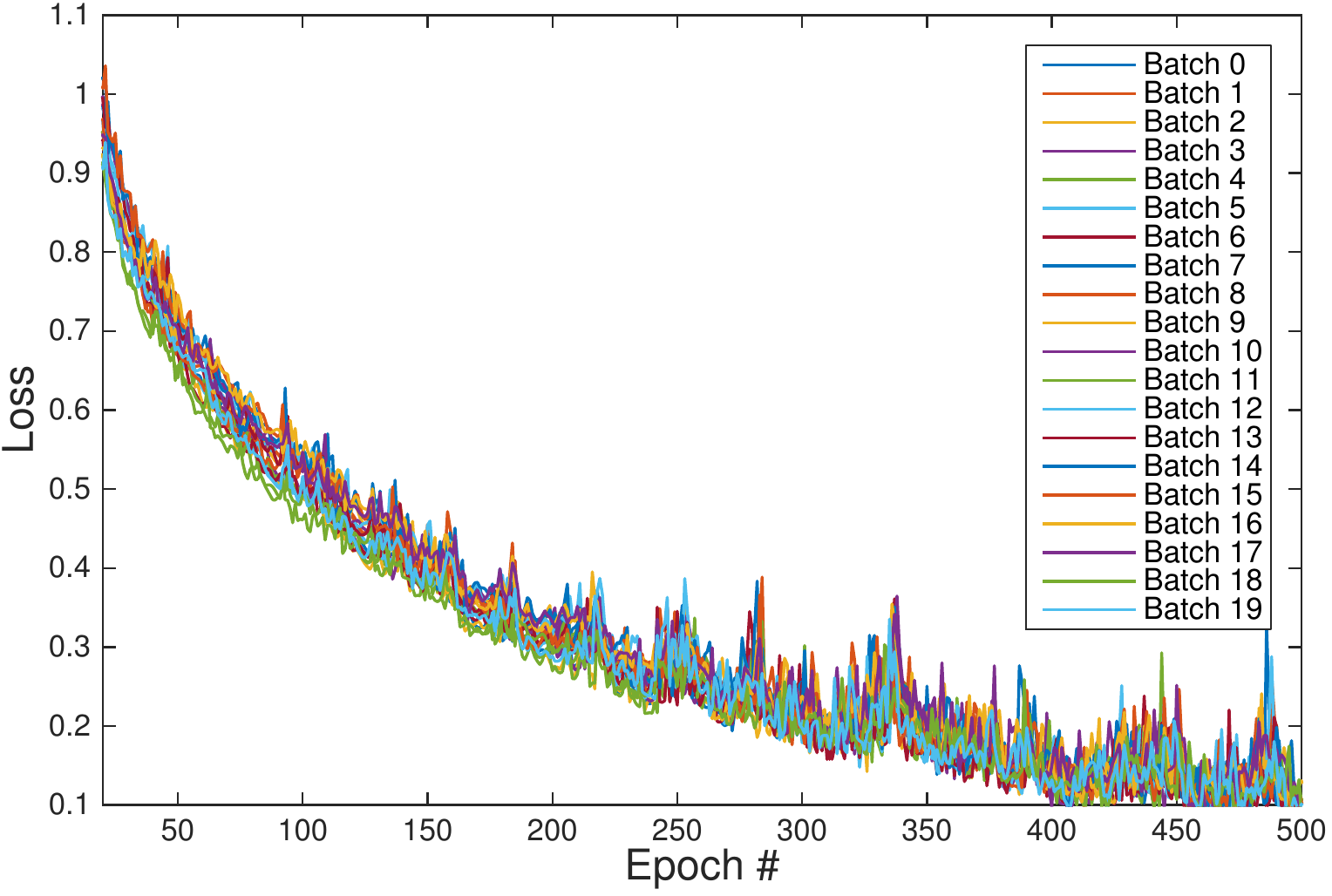}\label{isgd_losses}}
\subfloat[][SGD loss distribution by epochs]{\includegraphics[width=0.48\textwidth]{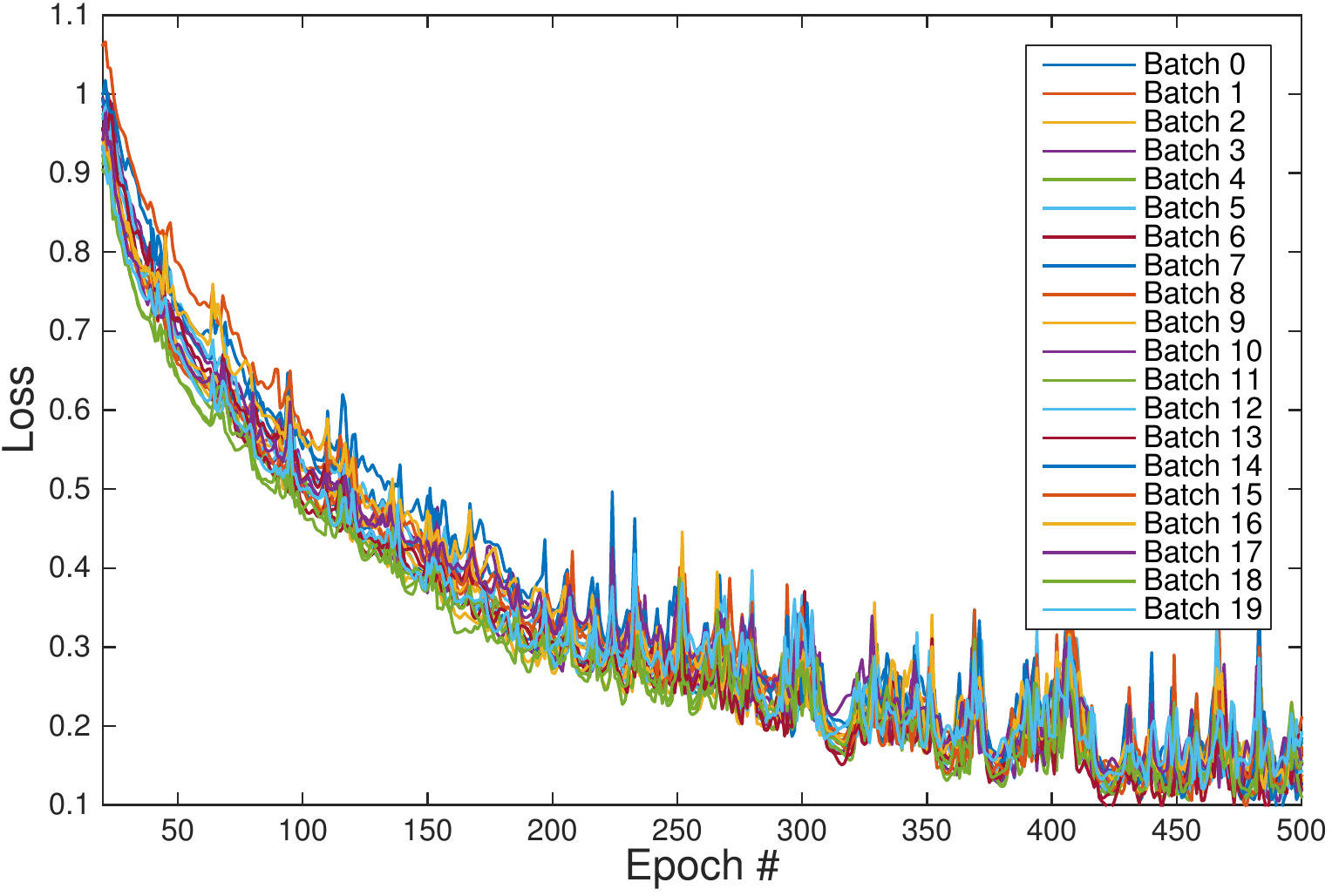}\label{sgd_losses}} \\
\subfloat[][STD of the batch's loss distribution]{\includegraphics[width=0.33\textwidth]{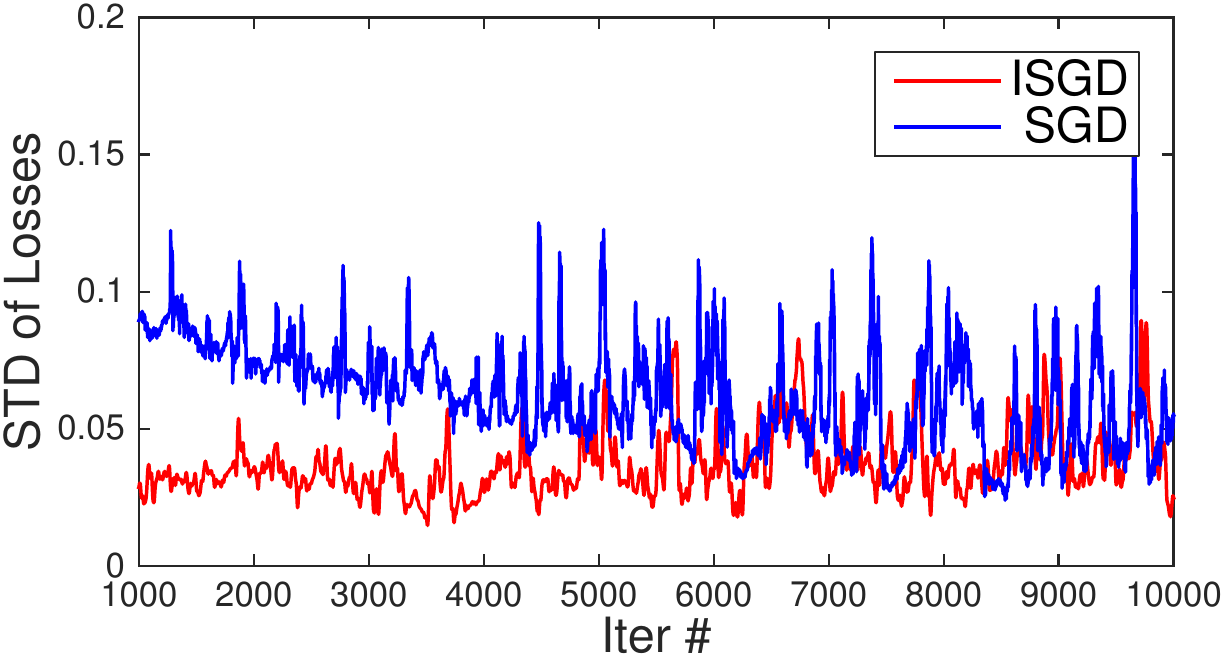}\label{loss_std}}
\subfloat[][average loss of 20 batches]{\includegraphics[width=0.33\textwidth]{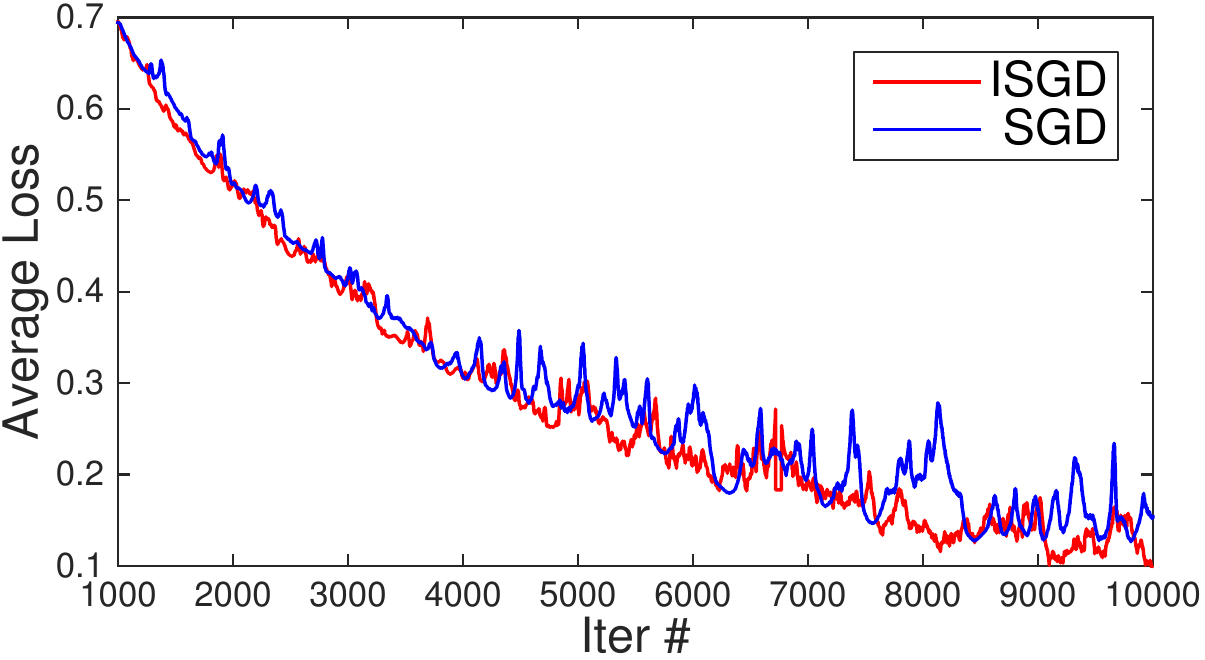}\label{loss_avg}}
\subfloat[][validation accuracy]{\includegraphics[width=0.33\textwidth]{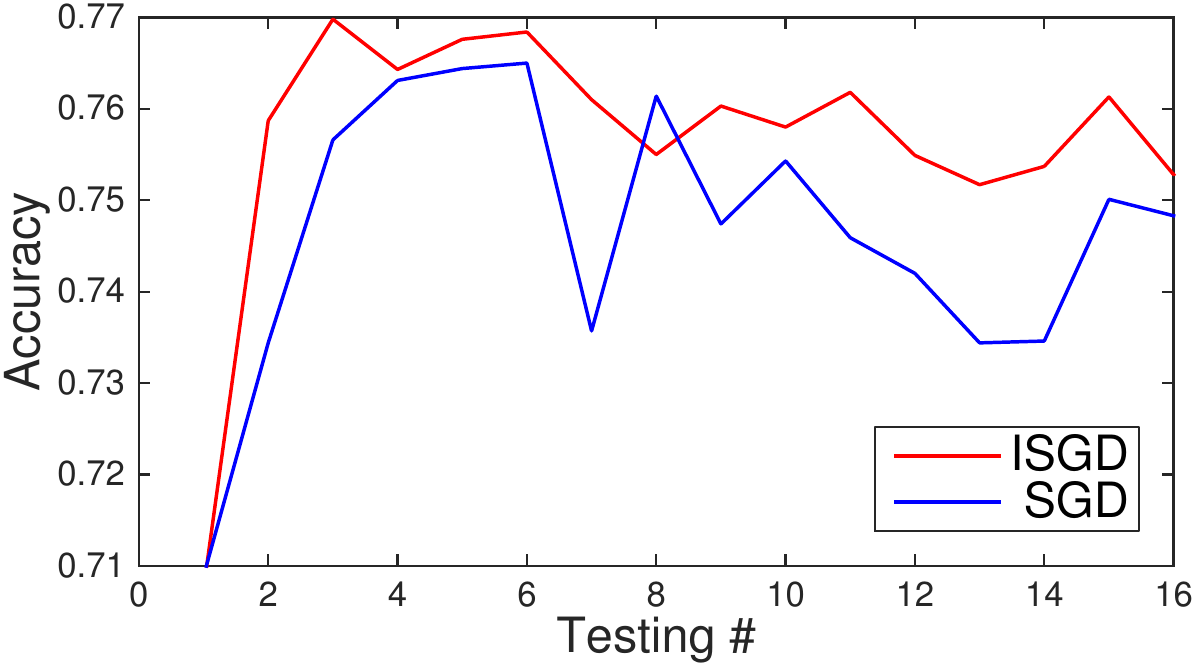}\label{loss_accuracy}}
\caption{The loss distribution, the average loss, the standard deviation of loss distribution, as well as the validation accuracy in the training on CIFAR.}
\label{loss_result}
\end{figure*}

\section{Experiments}
In this section, we demonstrate the performance of inconsistent training against SGD variants such as Momentum and Nesterov on a variety of widely recognized datasets including MNIST \cite{lecun1998mnist}, CIFAR-10 and ImageNet. MNIST has 60000 handwritten digits ranging from 0 to 9. CIFAR-10 has 60000 32$\times$32 RGB images categorized in 10 classes. ILSVRC 2012 ImageNet has 1431167 256 $\times$256 RGB images depicting 1000 object categories. We use LeNet, Caffe CIFAR-10 Quick, and AlexNet to train on MNIST, CIFAR-10, and ImageNet, respectively. The complexity of networks is proportional to the size of datasets. Therefore, our benchmarks cover the small, middle, and large scale CNN training.

We conduct the experiments on a multiGPU system with 4 NVIDIA Maxwell TITAN X. The CUDA version is 7.5, 
the compiler is GCC 4.8.4. The machine has 64 GB RAM and 1TB SSD. CPU is Xeon E5 4655 v3. Caffe is built with the cuDNN version 4. The GPU machine was exclusively owned by us during the benchmark.

\begin{figure*}[t]
\vspace{-0.3in}
\centering
\subfloat[][MNIST Test Accuracy]{\includegraphics[width=0.33\textwidth]{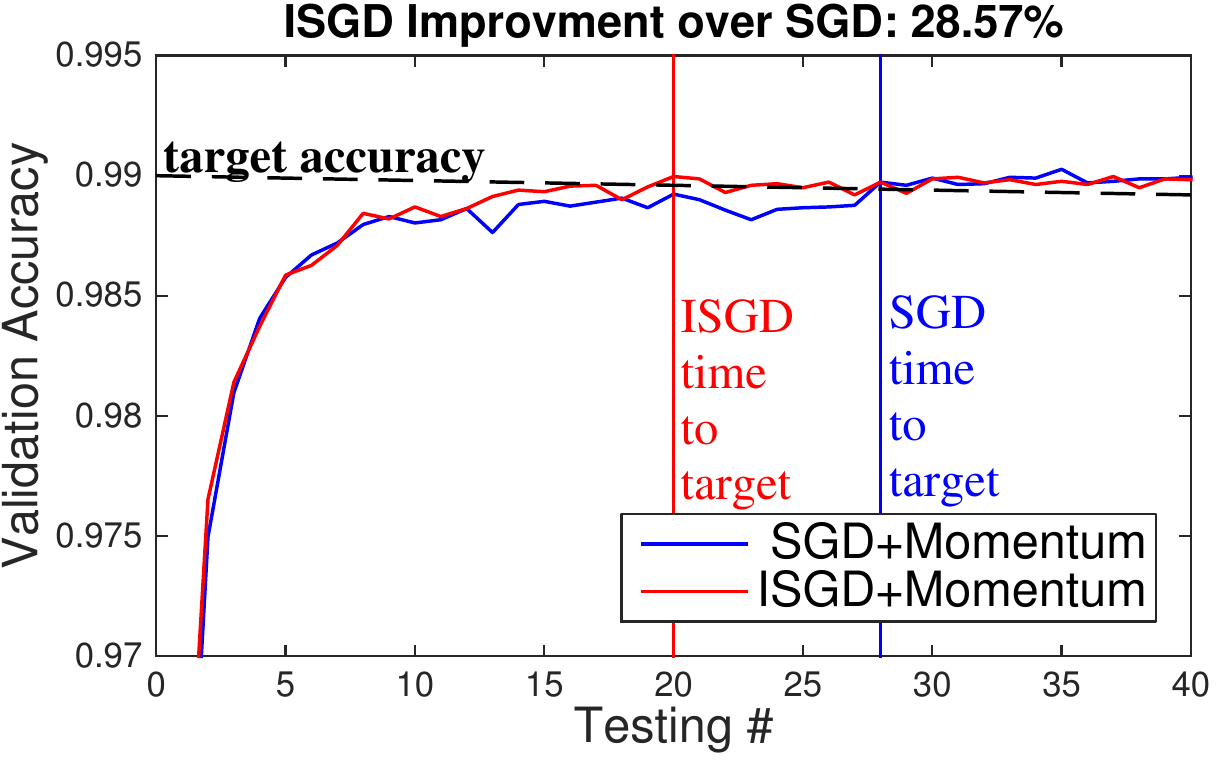}\label{MNIST_val}}
\subfloat[][CIFAR Test Accuracy]{\includegraphics[width=0.33\textwidth]{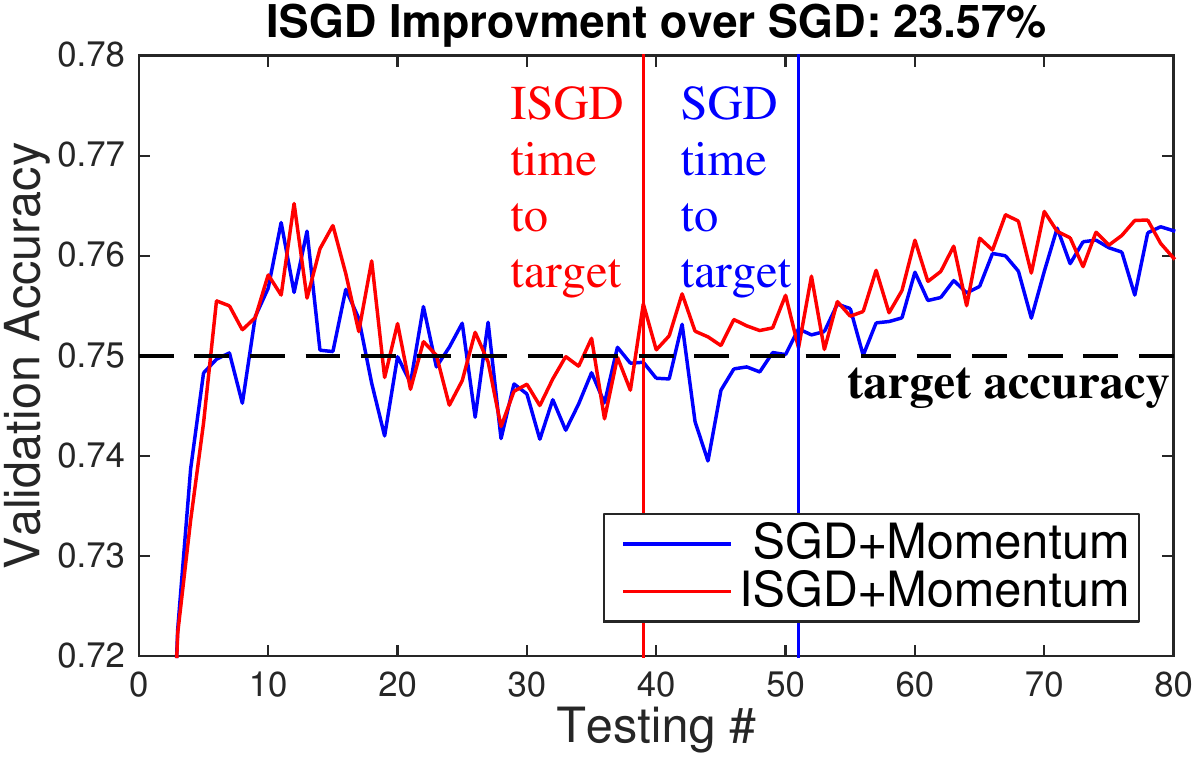}\label{CIFAR_val}}
\subfloat[][ImageNet Top 5 Accuracy]{\includegraphics[width=0.33\textwidth]{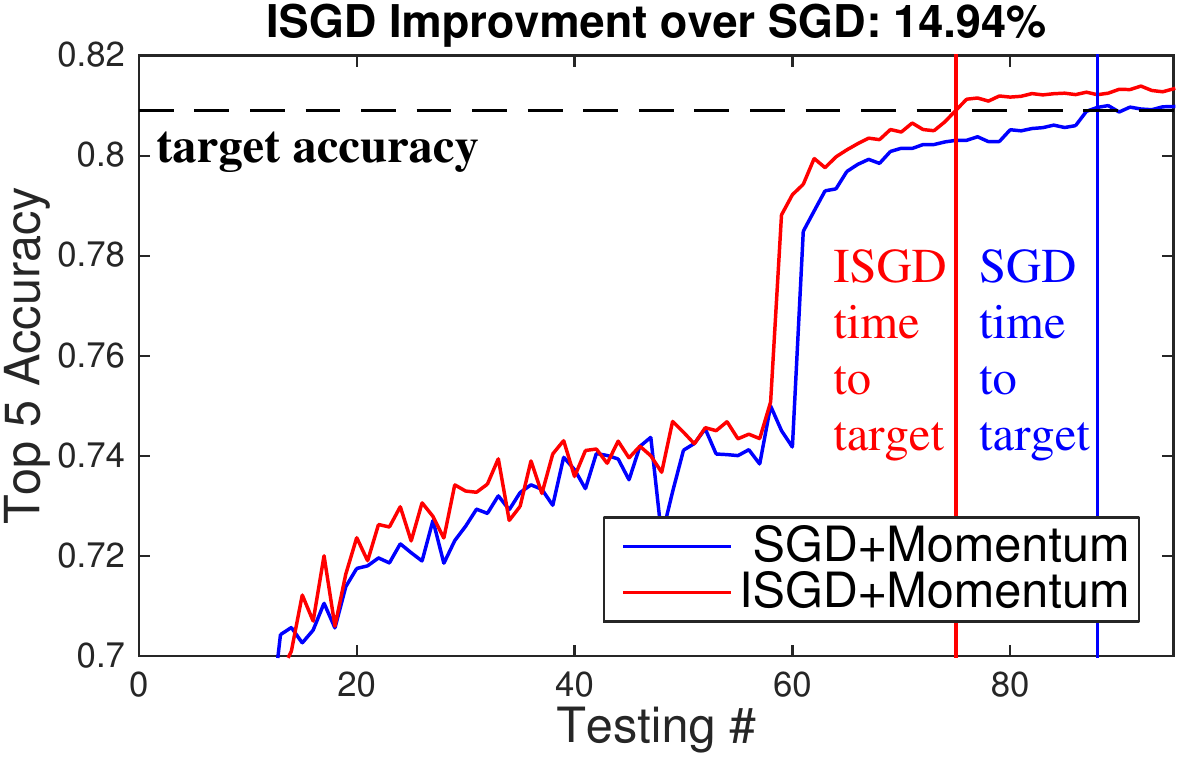}\label{imagenet_val_top5}} \\
\subfloat[][MNIST Train Error]{\includegraphics[width=0.33\textwidth]{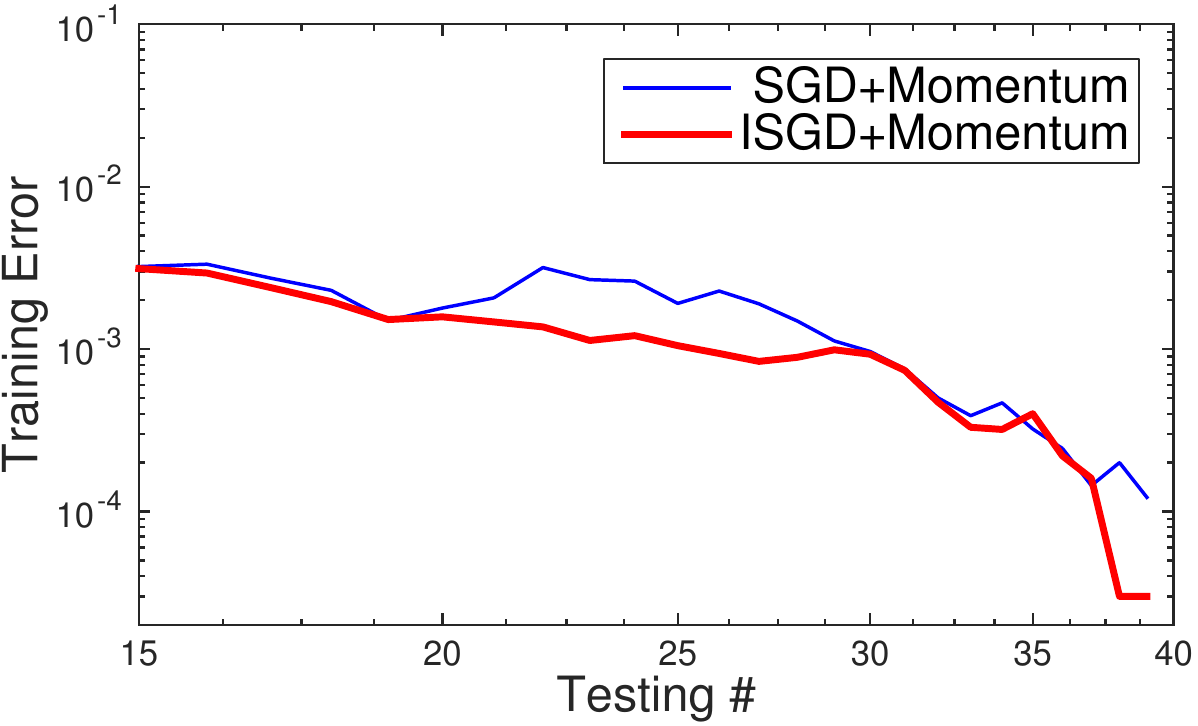}\label{MNIST_train}}
\subfloat[][CIFAR Train Error]{\includegraphics[width=0.33\textwidth]{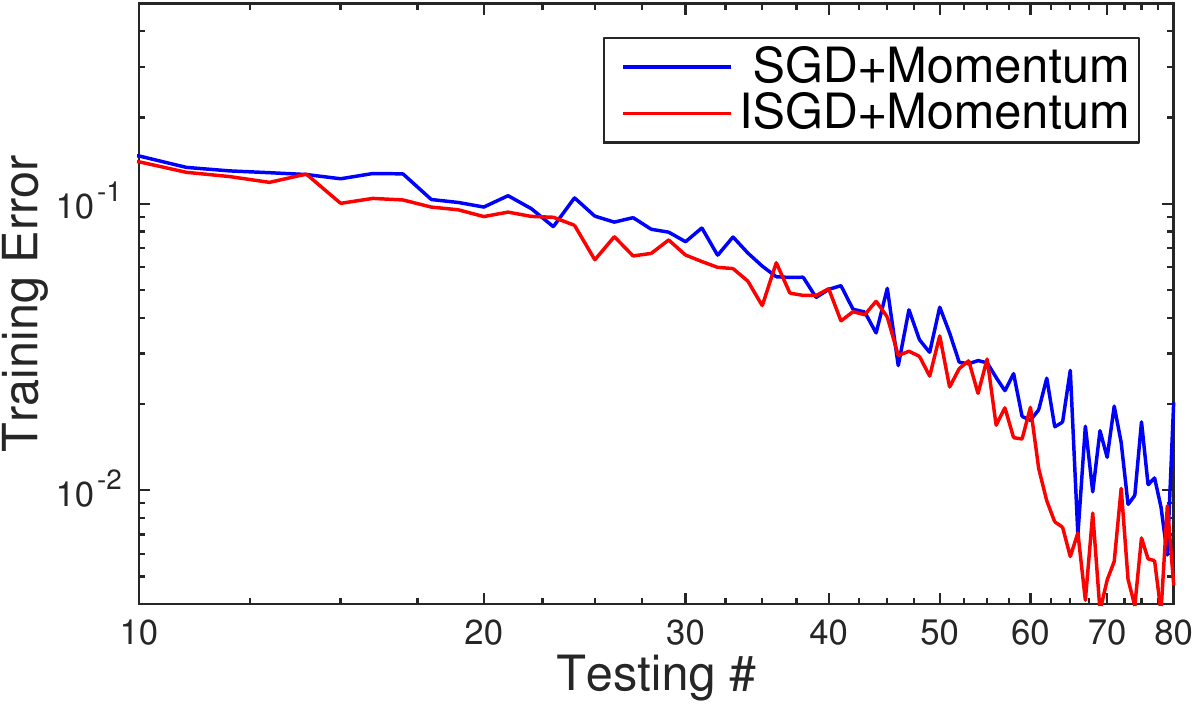}\label{CIFAR_train}}
\subfloat[][ImageNet Train Error]{\includegraphics[width=0.33\textwidth]{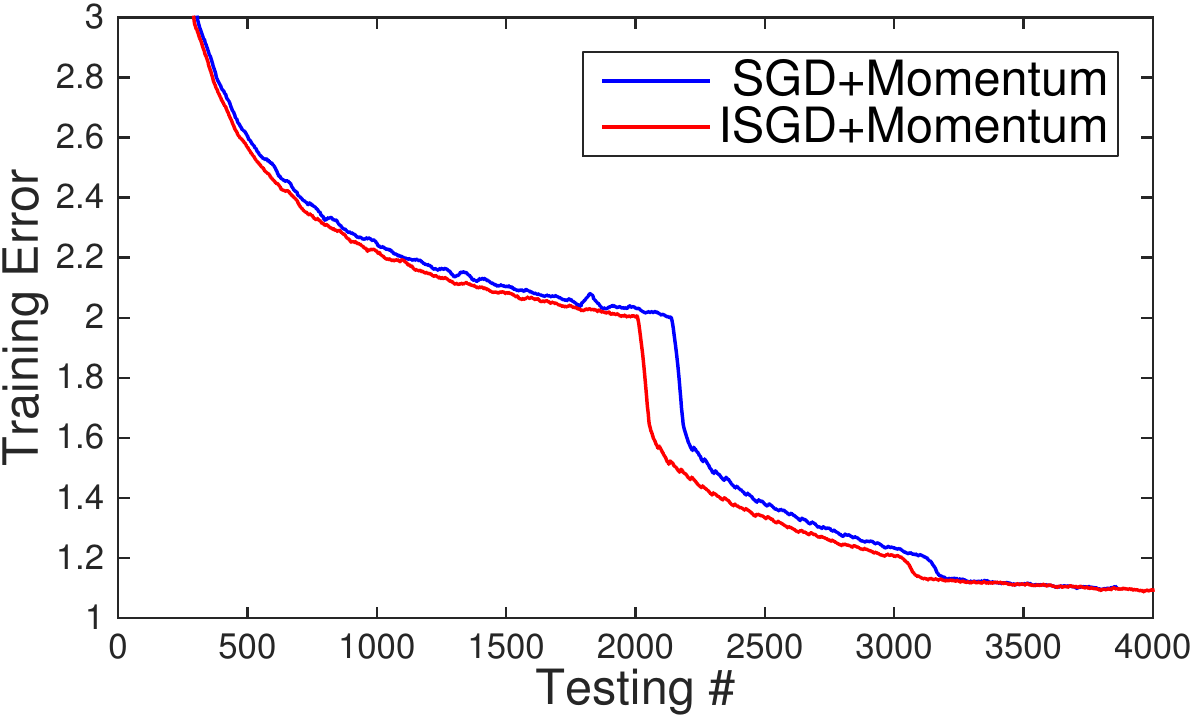}\label{imagenet_train_loss}}
\caption{The validation accuracy and training loss of LeNet, Caffe-Quick, AlexNet on MNIST, CIFAR and ImageNet.
ISGD consistently outperforms SGD.}
\label{test_result}
\end{figure*}

\begin{table*}[t]
  \scriptsize
    \centering
  \caption{ The average top accuracy and time reported from the training with ISGD and SGD on MNIST, CIFAR and ImageNet. IMP stands for the ISGD's improvement over SGD. The data is collected from 10 runs, and ISGD consistently outperforms SGD on 3 datasets.}
  \label{bounds}
  \begin{tabular}{ l l l l l l l l l l}
    \toprule
                  & \multicolumn{3}{c}{ Highest Reported Top/Top-5 Accuracy} & \multicolumn{3}{c}{ Average Top/Top-5 Accuracy} & \multicolumn{3}{c}{ Normalized Average Time to Top/Top-5 Accuracy. } \\
                  & SGD & ISGD &IMP & SGD & ISGD & IMP & SGD & ISGD & IMP\\
    \midrule
    \textbf{MNIST}        & $99.08\%$  &  $99.19\%$  & $0.11\%$  & $99.05\%$ & $99.17\%$ & $0.12\%  $ & 1 & 0.744 &  $ 25.6\%$  \\
    \textbf{CIFAR}         & $76.01\%$  & $76.55\%$ &   $0.54\%$ & $75.78\%$ & $76.42\% $ & $ 0.64\% $  & 1  & 0.772 &  $ 22.78\%$  \\
    \textbf{ImageNet }   & $82.12\%$  & $81.10\%$  &  $1.01\%$ & $81.83\%$ & $80.56\%$  & $1.27\%  $ & 1 &  0.8547 & 14.53\% \\
    \bottomrule
    \label{performance_data}
  \end{tabular}
\end{table*}

\subsection{Qualitative Evaluation of Inconsistent Training}
This section intends to qualitatively evaluate the impacts of inconsistent training. The purpose of inconsistent training is to rebalance the training effort across batches so that the large-loss batch receives more training than the small-loss one. To qualitatively evaluate the impacts of inconsistent training, we exam the progression of the loss distribution, the average loss, the standard deviation of batch's loss distribution, as well as the validation accuracy. We setup the training with Caffe CIFAR-10 Quick network on CIFAR-10 dataset. The batch size is set at 2500 yielding 20 independent batches. Fig.\ref{isgd_losses} and Fig.\ref{sgd_losses} present the loss distribution of 20 batches in the training. We arrange losses in epochs as the solver explores a batch only once in an epoch,

The inconsistent training has the following merits. 1) ISGD converges faster than SGD due to the improvement of training model. We measure the convergence rate by the average loss of batches in a dataset, and the method conforms to the training definition in Eq.\ref{sgd_objectives}. The average loss data in Fig.\ref{loss_avg} demonstrates that ISGD converges faster than SGD. In contrast with SGD, the lower average loss of ISGD after $iter > 7000$ (Fig.\ref{loss_avg}) explains the better accuracy of ISGD after testing 9 (Fig.\ref{loss_accuracy}). The validation accuracy of ISGD in Fig.\ref{loss_accuracy} is also above SGD, that is consistent with data in Fig.\ref{loss_avg} that the average loss of ISGD is below the average loss of SGD in the training. These justify the convergence advantage of inconsistent training. 2) ISGD dynamically accelerates the large-loss batch in the progress of training to reduce the training gap with small-loss batches. Therefore, the variation of batch's training status is less than the one trained by SGD. Please note we measure the training status of a batch with its loss, and the variation of batch's training status is measured by the standard deviation of batch's loss distribution. Fig.\ref{loss_std} demonstrates the inconsistent training successfully attenuates the training variations among batches. When $iter \in [1000, 6000]$, the std of batch's loss distribution of ISGD is much lower than SGD. The result is also consistent with the loss distribution in Fig.\ref{isgd_losses} and Fig.\ref{sgd_losses}, in which the loss distribution of SGD is much wider than ISGD at $epoch \in [50, 300]$.

\begin{figure*}[t]
\centering
\subfloat[][MNIST]{
\includegraphics[width=0.33\textwidth]{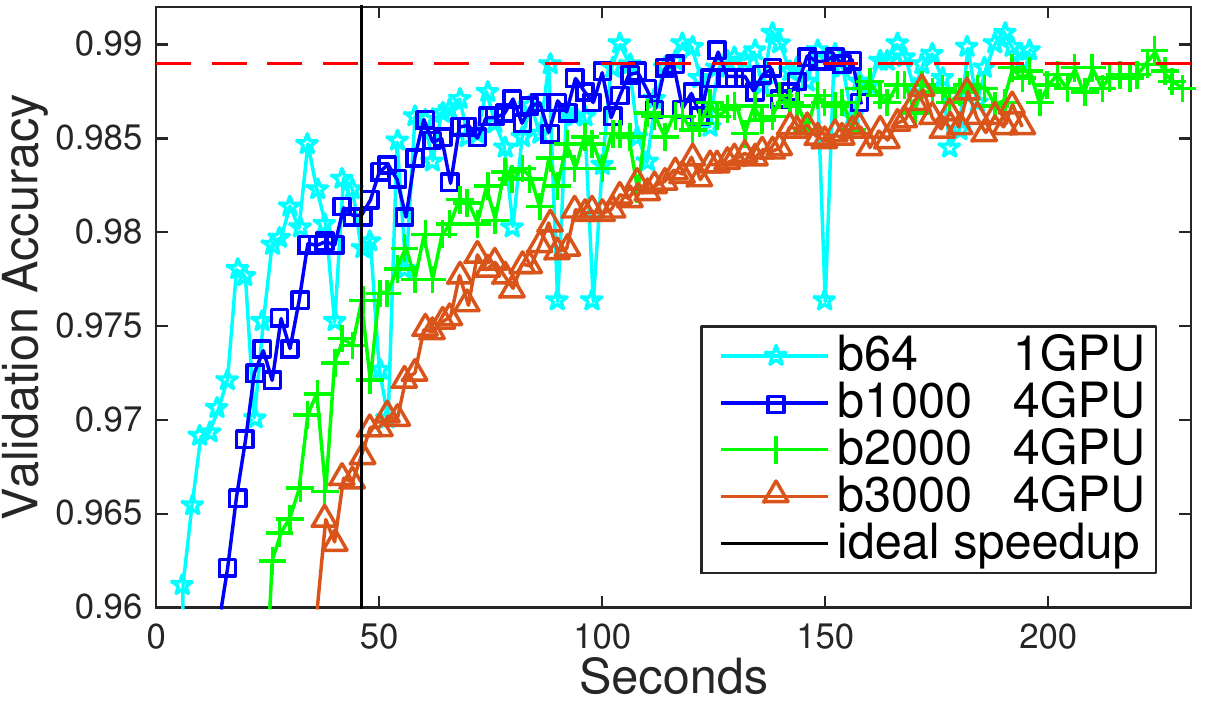}\label{MNIST_scal}
}
\subfloat[][CIFAR]{
\includegraphics[width=0.33\textwidth]{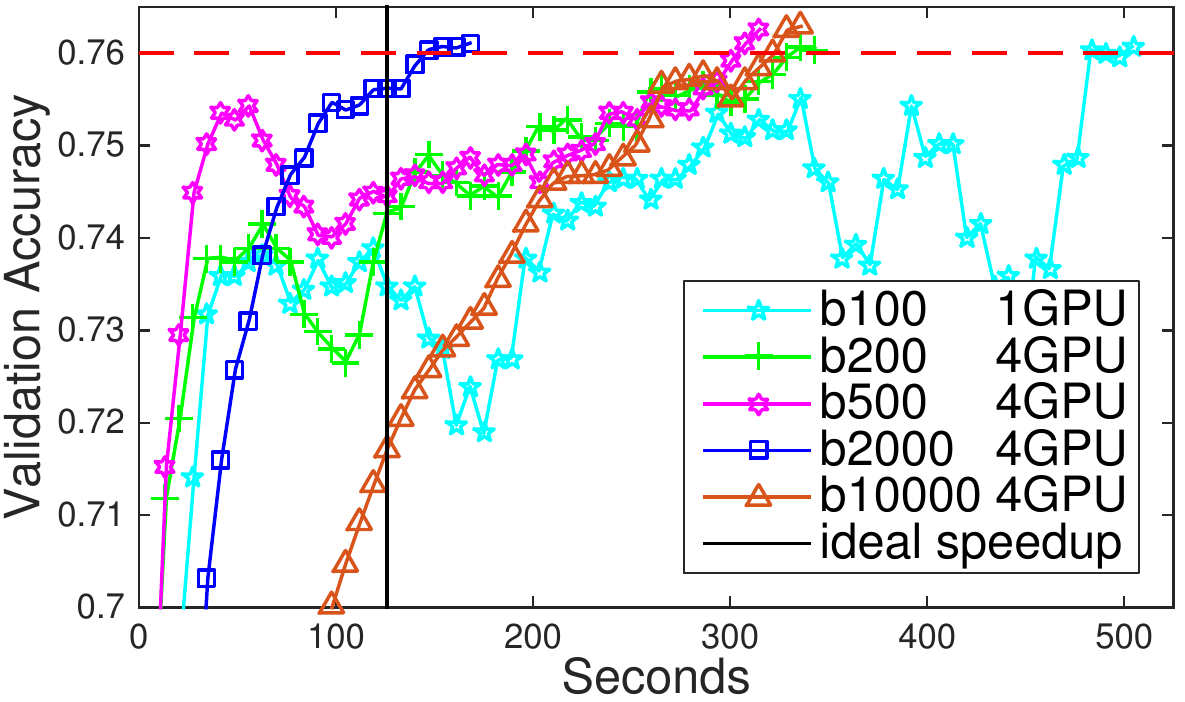}\label{CIFAR_scal}
}
\subfloat[][ImageNet]{
\includegraphics[width=0.33\textwidth]{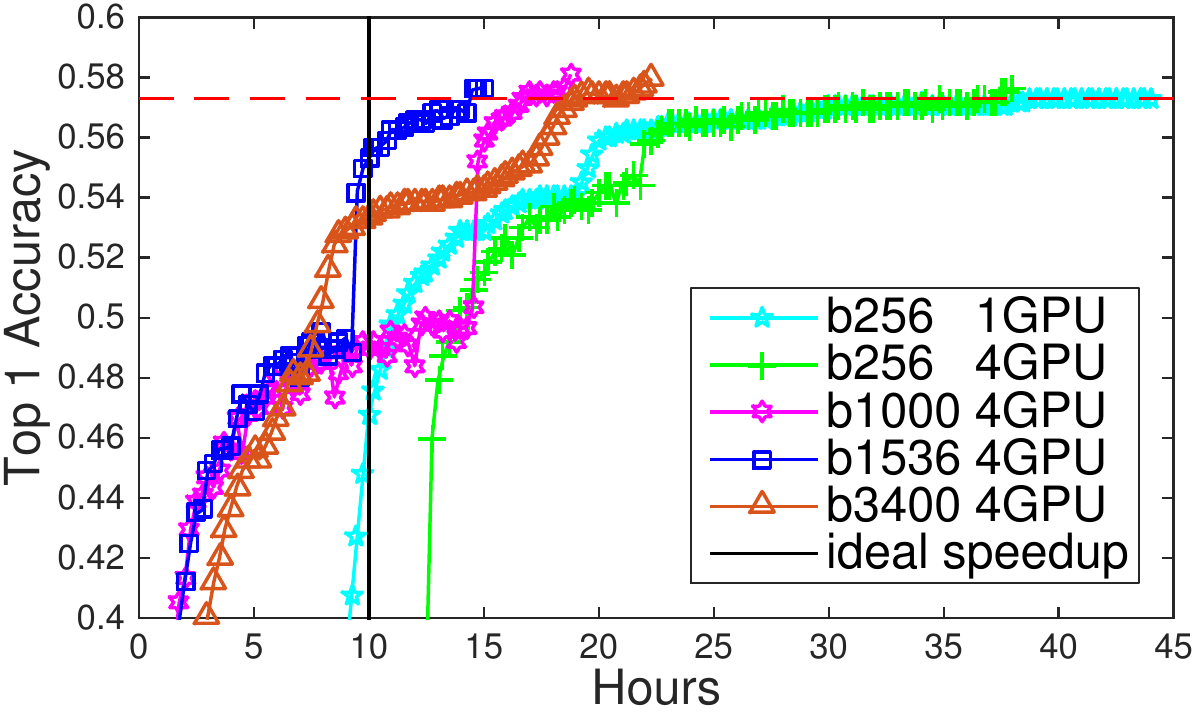}\label{imagenet_scal}
}
\caption{The effect of batch size on the total training time. The experiments are conducted with 4 TITAN-X GPUs. }
\label{scalability}
\end{figure*}

\begin{figure}[t]
\centering
\subfloat[][ImageNet Train Error]{
\includegraphics[width=0.33\textwidth]{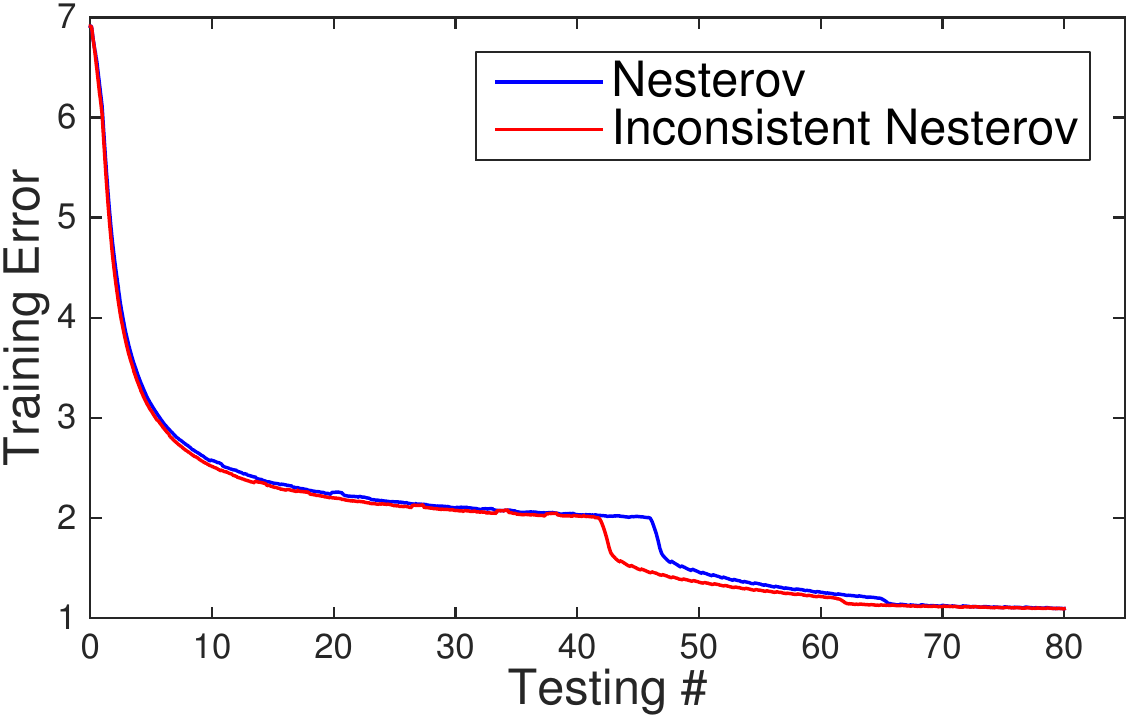}\label{ines_train}
} \\
\subfloat[][ImageNet Top 1 Accuracy]{
\includegraphics[width=0.33\textwidth]{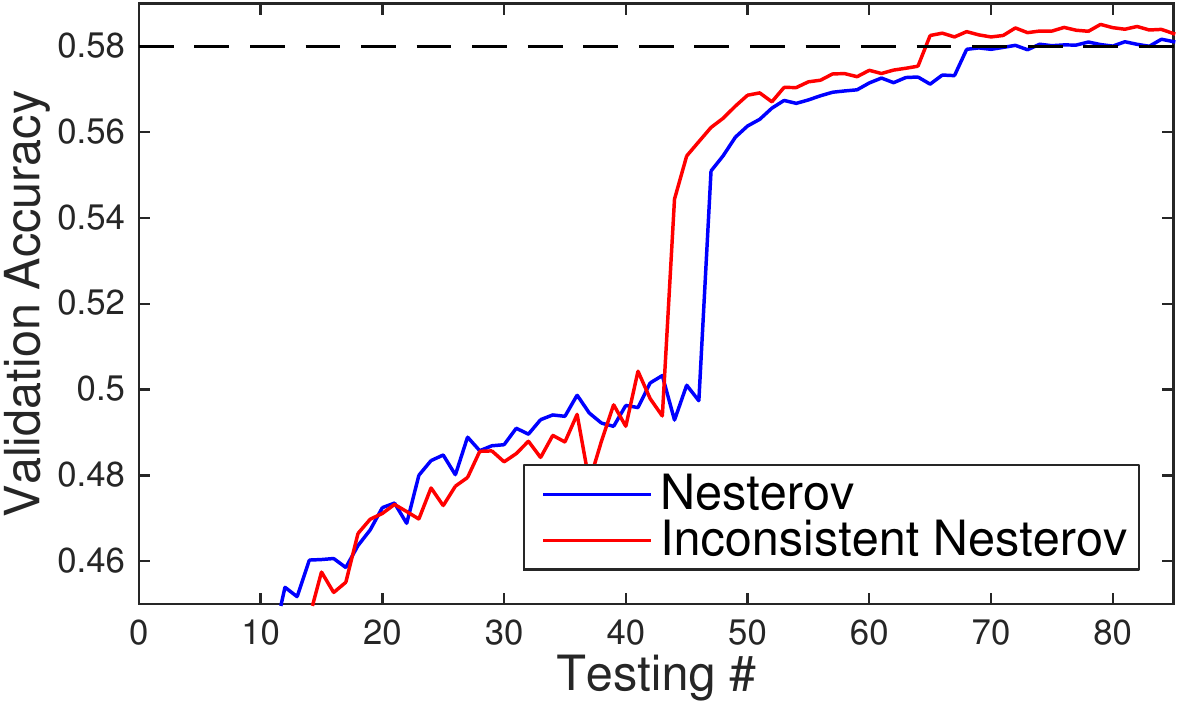}\label{ines_val}
}
\caption{Train ImageNet with the inconsistent Nesterov accelerated gradient.}
\label{inconsistent_nes}
\end{figure}

\subsection{Performance Evaluations}
The setup of each comparisons, ISGD V.S. SGD, has been carefully set to be the single factor experiment, i.e. the only difference is the inconsistent training. Some parameters of SGD greatly affect the training performance, setting different values on them jeopardizes the credibility of experiments. Therefore, we ensure the parameters of SGD and ISGD to be same in each comparisions. The first parameter considered is the learning rate. The MNIST tests adopt a constant learning rate of 0.01, and CIFAR tests adopt a constant learning rate of 0.001. Both cases are consistent with the solver defined in Caffe. Caffe fixes the learning rate for these two cases because networks yield the satisfactory accuracies , $75\%$ on CIFAR and $99\%$ on MNIST, without shrinking the learning rate. Since AlexNet has to shrink $lr$, the learning rate of it has 3 possibilities: lr = 0.015 if the average loss $\overline{\psi} \in [2.0, +\infty]$, lr = 0.0015 if $\overline{\psi}$ in $[1.2, 2.0)$, and lr = 0.00015 if $\overline{\psi}$ in $[0, 1.2)$. The batch size is also same for each comparison in CIFAR, MNIST and ImageNet. We adopt a large batch to fully saturate 4 GPUs. For other parameters such as the weight decay and momentum, they are also same through all the tests. 

ISGD consistently outperforms SGD in all tests manifesting the effectiveness of inconsistent training. Please note both methods incorporate the momentum term. Since an iteration of ISGD is inconsistent, we test every other 2, 6, 900 seconds (only count the training time with the test time excluded) for MNIST, CIFAR and ImageNet tests, respectively. The horizontal dashed line represents the target accuracy, and the total training time starts from 0 to the point that the validation accuracy is consistently above the dashed line. In the ImageNet test, ISGD demonstrates the 14.94\% faster convergence than SGD. SGD takes 21.4 hours to reach the 81\% top 5 accuracy, while ISGD takes 18.2 hours (Fig.\ref{imagenet_val_top5}) . In the CIFAR test, ISGD demonstrates 23.57\% faster convergence than SGD. The top accuracy for CIFAR-Quick network reported on CIFAR-10 is 75\%. After 306 seconds, the test accuracy of SGD is steadily above 75\%, while ISGD only takes 234 seconds  (Fig.\ref{CIFAR_val}). Finally, ISGD demonstrates 28.57\% faster convergence than SGD on MNIST dataset. It takes SGD 56 seconds to reach the 99\% top accuracy, while ISGD only takes 40 seconds. Since the training is essentially a stochastic process, the performance subjects to changes. We repeat each test cases 10 times, and we list the performance data in Table.\ref{performance_data}. The results also uphold the convergence advantages of inconsistent training.

To explain the performance advantages of ISGD, we also use the training dataset to test. Whereas, the training set of ImageNet 256 GB is too large to be tested, we use $\overline{\psi}$ in Alg.\ref{ugd_alg} to approximate the training error. Fig.\ref{MNIST_train}, Fig.\ref{CIFAR_train} and Fig.\ref{imagenet_train_loss} demonstrate the training error of ISGD is consistently below the SGD. The results demonstrate the benefit of inconsistent training, and they also explain the good validation accuracy of ISGD in Fig.\ref{MNIST_val}, Fig.\ref{CIFAR_val} and Fig.\ref{imagenet_val_top5}.

The inconsistent training is also compatible with the Nesterov accelerated gradient. Fig.\ref{inconsistent_nes} demonstrates the validation accuracy and the training loss progression on ImageNet trained with the Nesterov accelerated gradient. The inconsistent training beats the regular Nesterov method. If set 58\% top 1 accuracy as the threshold, the inconsistent training takes 65 tests to exceed the threshold, while the regular one takes 75 tests. Please note the time interval of two consecutive tests is fixed. Therefore, the inconsistent training demonstrates the 13.4 \% performance gain. The compatibility is under our expectation. The Nesterov method accelerates the convergence by considering the curvature information, while ISGD rebalances the training across batches.

\subsection{Time Domain Convergence Rate W.R.T Batch Size on MultiGPUs}
Fig.\ref{scalability} demonstrates convergence speeds at different batch sizes on MNIST, CIFAR and ImageNet datasets. The figures reflect the following conclusions: 1) A sufficiently large batch is necessary to the multiGPU training. The single GPU only involves computations $t_{compt}$, while the multiGPU training entails an additional term $t_{comm}$ for synchronizations. A small batch size for the single GPU training is favored to ensure the frequent gradient updates. In the multiGPU training, the cost of synchronizations linearly increases with the number of gradient updates. Increasing batch size improves the convergence rate, thereby fewer iterations and synchronizations. Besides, it also improves system utilizations and saturations. As a consequence, a moderate batch size is favored to the multiGPU training as indicated in Fig.\ref{scalability}. 2) An unwieldy batch size slows down the convergence. Because computations linearly increase with the batch size, which reduces the number of gradient updates in a limited time. The declined convergence speed is observable in the Fig.\ref{MNIST_scal}, Fig.\ref{CIFAR_scal} and Fig.\ref{imagenet_scal} when batch size is set at 3000, 10000, 3400, respectively.

\section{Summary}

In this paper, we propose the inconsistent training to dynamically adjust the training effort w.r.t batch's training status. ISGD models the training as a stochastic process, and it utilizes techniques in Stochastic Process Control to identify a large-loss batch on the fly. Then, ISGD solves a new subproblem to accelerate the training on the under-trained batch. Extensive experiments on a variety of datasets and models demonstrate the promising performance of inconsistent training.

\newpage
\renewcommand\refname{Bibliography}
\bibliographystyle{elsarticle-num}
\bibliography{ref}

\end{document}